\pdfoutput=1
\documentclass[sigconf,nonacm]{acmart}
\usepackage{graphicx} 
\graphicspath{{images/}}
\usepackage{booktabs} 
\usepackage{array} 
\usepackage{caption} 
\usepackage[title]{appendix}
\usepackage{algorithm}
\usepackage{algpseudocode}
\usepackage{amsmath}
\usepackage{amsfonts}
\usepackage{multirow}
\usepackage{tablefootnote}
\usepackage{wrapfig}
\usepackage{arydshln}
\usepackage{url}
\usepackage{subcaption}
\usepackage{float}
\begin{document}
\title{FAST-Q: Fast-track Exploration with Adversarially Balanced State Representations for Counterfactual Action Estimation in Offline Reinforcement Learning}

\author{Pulkit Agrawal}
\email{pulkit.agrawal@games24x7.com}
\affiliation{%
  \institution{Artificial Intelligence and Data Science, Games24x7}
  \country{India}
}
\author{Rukma Talwadker}
\email{rukma.talwadker@games24x7.com}
\affiliation{%
  \institution{Artificial Intelligence and Data Science, Games24x7}
  \country{India}
}
\author{Aditya Pareek}
\email{aditya.pareek@games24x7.com}
\affiliation{%
  \institution{Artificial Intelligence and Data Science, Games24x7}
  \country{India}
}
\author{Tridib Mukherjee}
\email{tridib.mukherjee@games24x7.com}
\affiliation{%
  \institution{Artificial Intelligence and Data Science, Games24x7}
  \country{India}
}

\begin{abstract}
Recent advancements in state-of-the-art (SOTA) offline reinforcement learning (RL) have primarily focused on addressing function approximation errors, which contribute to the overestimation of Q-values for out-of-distribution actions — a challenge that static datasets exacerbate. However, high-stakes applications such as recommendation systems in online gaming, introduce further complexities due to players' psychology/ intent driven by gameplay experiences and the platform’s inherent volatility. These factors create highly sparse, partially overlapping state spaces across policies, further influenced by the \textit{experiment path} selection logic which biases state spaces towards specific policies. Current SOTA methods constrain learning from such offline data by clipping known counterfactual actions as out-of-distribution due to poor generalization across unobserved states. Further aggravating conservative Q-learning and necessitating more online exploration. \textbf{FAST-Q} introduces a novel approach that (1) leverages \textit{Gradient Reversal Learning} to construct \textbf{balanced state} representations, regularizing the policy-specific bias between the players’ state and action thereby enabling counterfactual estimation; (2) supports \textbf{offline counterfactual exploration} in parallel with static data exploitation; and (3) proposes a Q-value decomposition strategy for multi-objective optimization, facilitating \textbf{explainable recommendations} over short and long-term objectives. These innovations demonstrate superiority of FAST-Q over prior SOTA approaches and demonstrates \textit{at least} \textbf{0.15\%} increase in player returns, \textbf{2\%} improvement in lifetime value (LTV), \textbf{0.4\%} enhancement in the recommendation driven engagement, \textbf{2\%} improvement in the players' platform dwell time and an impressive \textbf{10\%} reduction in the costs associated with the recommendation, on our volatile gaming platform.
\end{abstract}

\maketitle

\section{Introduction}
\label{sec:intro}
Offline reinforcement learning (RL), or batch RL, has revolutionized recommender systems by effectively modeling dynamic long-term preferences and short-term stochastic patterns using offline data~\cite{lange,fujimoto2019}. 
In skill-based real-money games, such as Rummy~\cite{Rummy_Wikipedia}, player behavior exhibits significant stochasticity, driven by hidden intent shaped by past gameplay experiences ~\cite{ARGO,cognitionnet}.  For instance, a certain player in response to losses might become demotivated and hence dis-engaged, while another could get highly-motivated towards chasing wins - two extremities resulting from the same outcome. Decisions rooted in this organic intent can lead to varied effects: they might reduce immediate engagement but foster long-term retention, conversely, encourage overspending and increase the risk of churn.

Our objective is to recommend optimal challenges for each player, balancing immediate and long-term engagement while managing associated costs on our volatile platform. Challenges consist of a specific number of games and target scores, paired with cash rewards upon completion. 
On our platform, a range of recommendation policies were developed---(i) Policy-1: tailoring player targets based on each player's past performance; (ii) Policy-2: incorporating the player's intent to play on a given day~\cite{ARGO} to predict challenges; and (iii) Policy-3: relying solely on the player's organic engagement, offering no additional rewards beyond the wins on the table (i.e., no specific challenges). None of these policies consistently outperformed each other in our online A/B testing setup and catered to players' long-term engagement. 
This motivated us to build \textbf{FAST-Q} which was presented with some  unique challenges.\footnote{This is the author's version of the work accepted for publication in the  ACM WWW Companion'25. The final version is published by ACM and is available at:
https://doi.org/10.1145/3701716.3715224}

\begin{figure}
\begin{minipage}{0.68\linewidth}
\includegraphics[width=\linewidth]{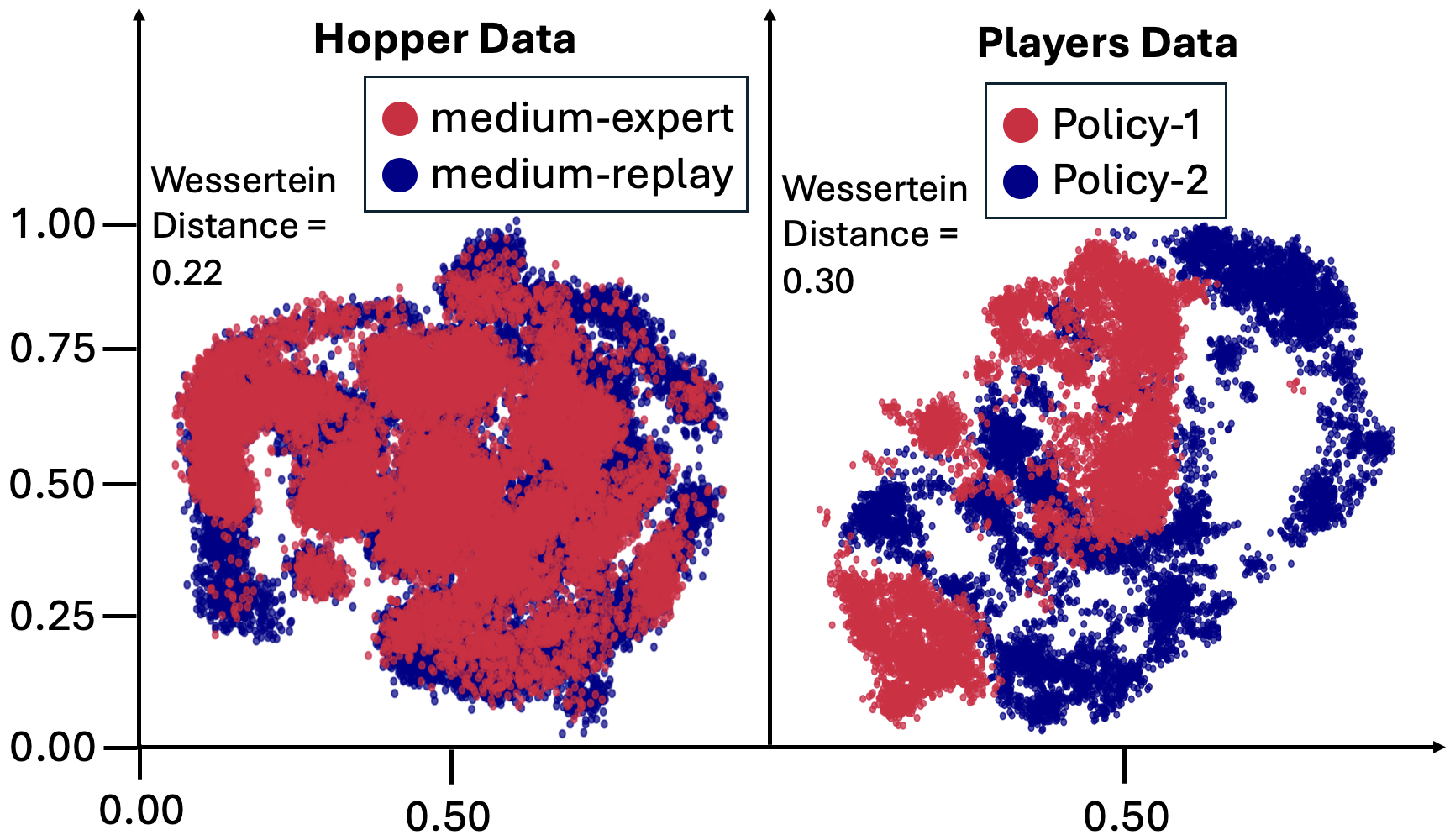}
\caption{\textmd{Difference in state isolation across policies in Gym-MuJoCo dataset w.r.t Players Data}}
\label{fig_state_isolation_across_policy_classes}
\end{minipage}
\hfill
\begin{minipage}{0.30\linewidth}
\includegraphics[width=\linewidth]{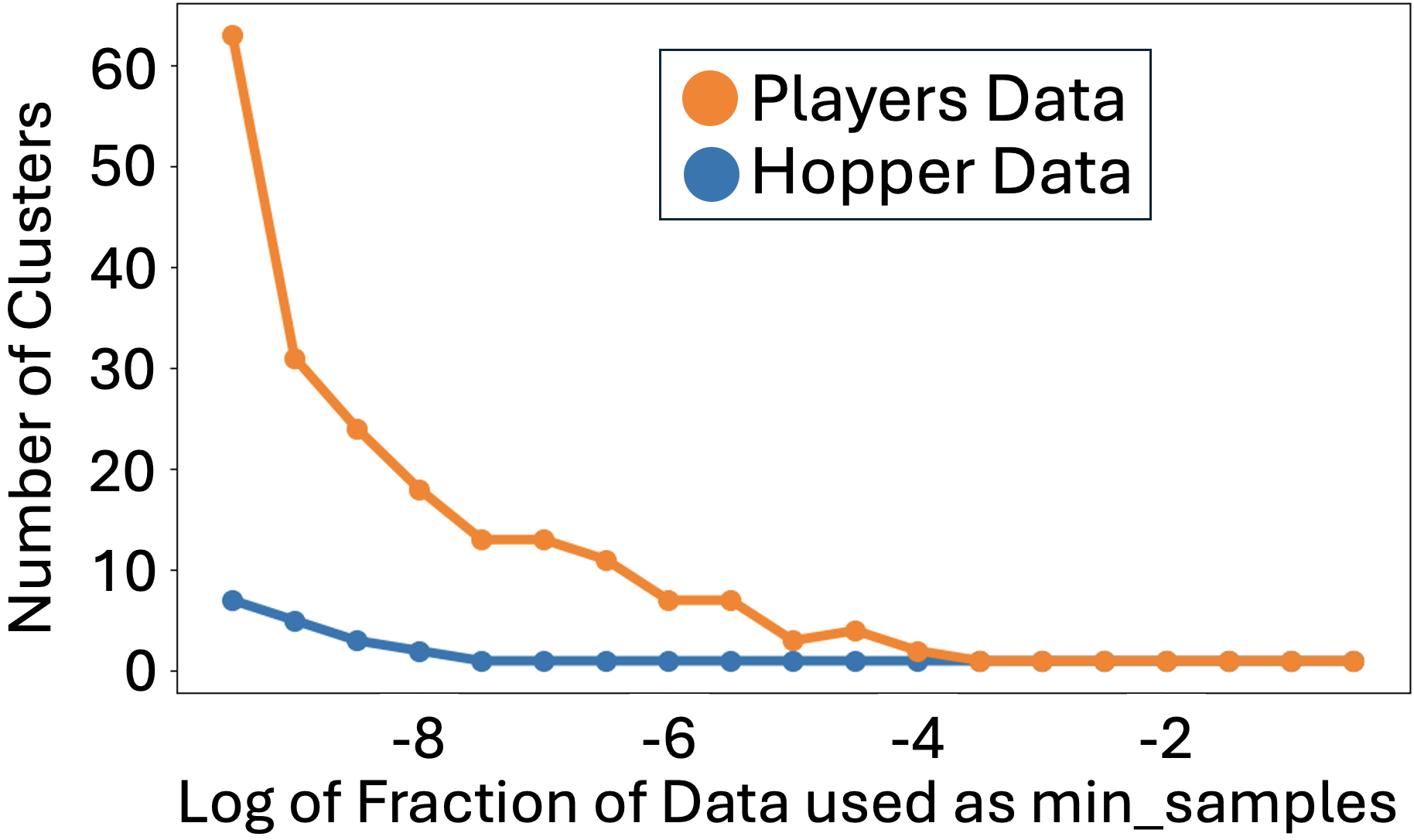}
\caption{\textmd{Higher data sparsity in our Players Data}}
\label{fig_state_isolation_sparsity}
\end{minipage}
\end{figure}

\textbf{1. Variable Action Adaptations over Disparate State Spaces:} Figure \ref{fig_state_isolation_across_policy_classes} illustrates more disparate and distinct assignment of state spaces across two policy classes on our platform with higher sum of the Wasserstein distance across the two dimensions, compared to the mixed-policy Gym-MuJoCo Hopper task in the D4RL dataset ~\cite{fu2021d4rl} used as a representative benchmark. Figure \ref{fig_state_isolation_sparsity} also illustrates higher degree of sparsity in our data, with our data consistently exhibiting a greater number of distinct clusters, indicating more sparse data distribution. Further details on the methods used is in Appendix~\ref{appendix:sparsity}. Figure \ref{fig:varying_actions_across_policies} further highlights the disparity in recommended actions across policy classes when provided with identical state representations, showing a higher mean on the difference between actions as compared to the Hopper dataset. Policy switch line in the Figure \ref{fig:varying_actions_across_policies} to be discussed in the Section \ref{sec:offline_obj}

Exclusivity in state-action representations stemmed from various factors, including the independent evolution of policy classes and the differing platform priorities shaped by the nature of player engagement across these paths. 




\begin{figure}[htp]
\subcaptionbox{a) Hopper Data \label{subfig:fig1}}{
  \includegraphics[width=0.45\textwidth, height=3cm]{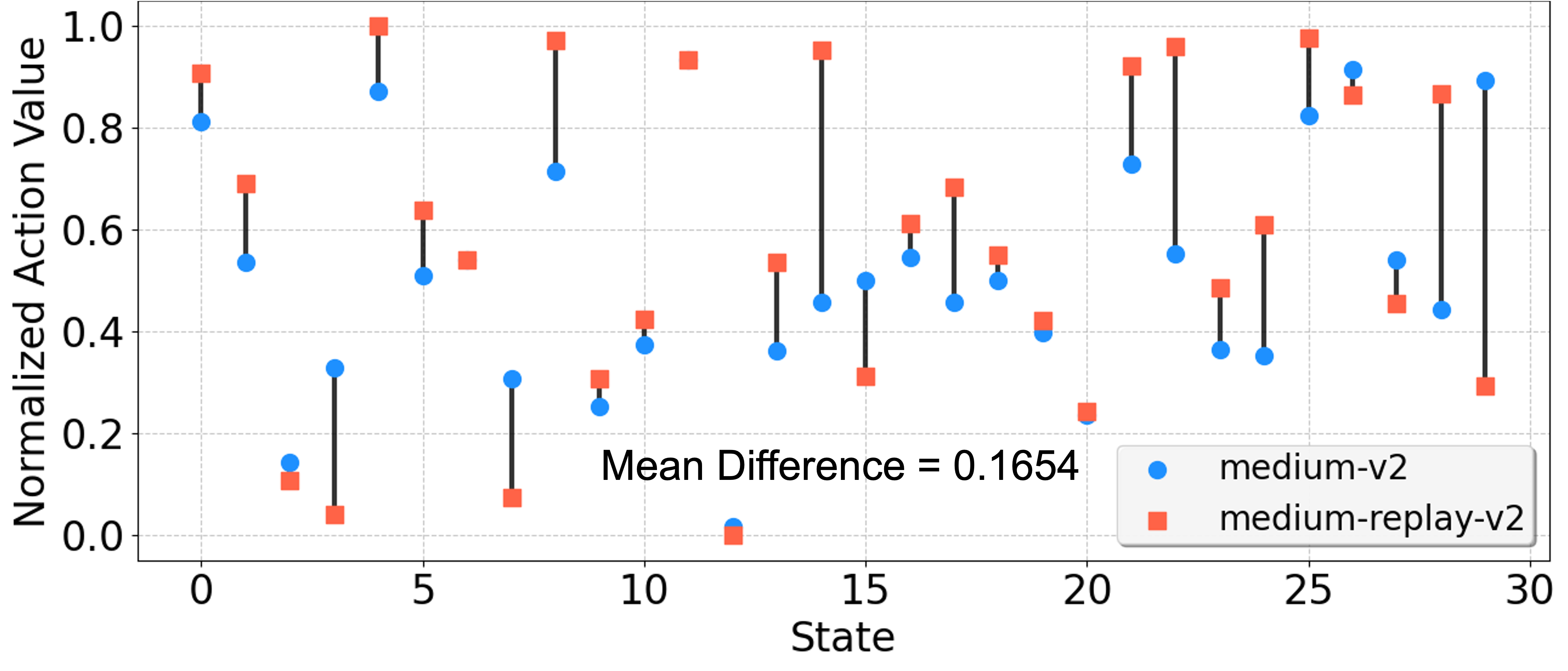}
}
\vspace{1ex}

\subcaptionbox{b) Player Data \label{subfig:fig2}}{
  \includegraphics[width=0.45\textwidth, height=3cm]{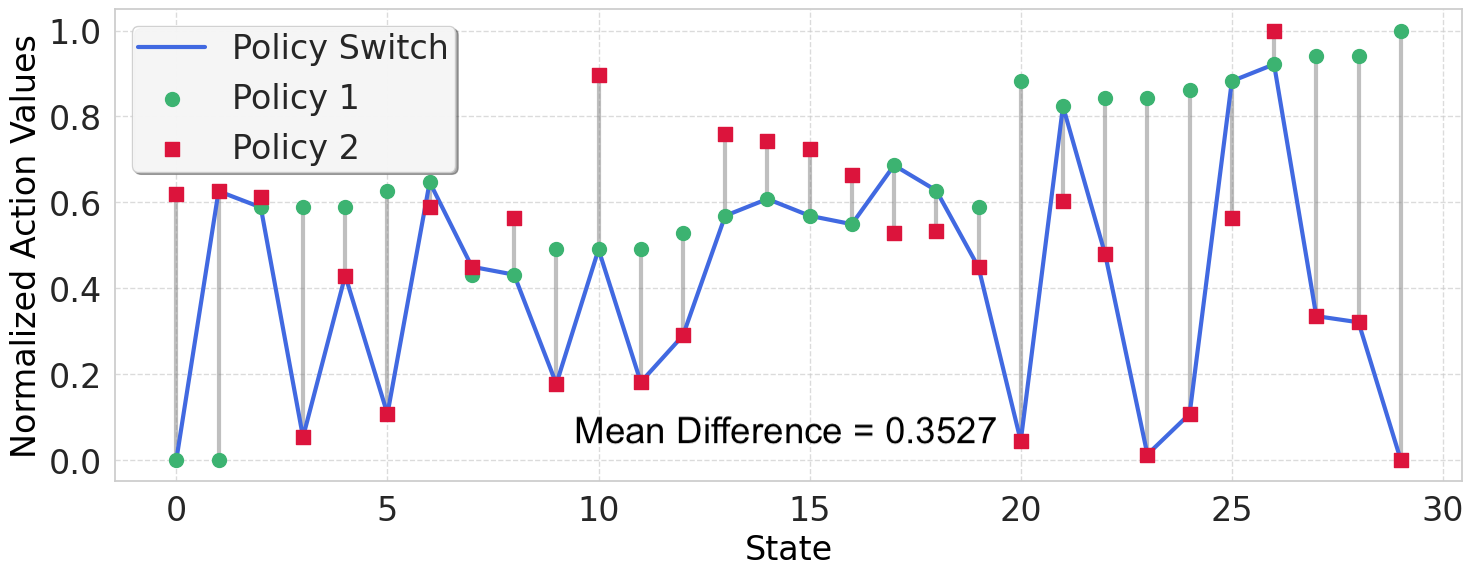}
}
\vspace{-0.1in}
\caption{\textmd{High Variability of actions across policies for identical states on our platform compared to Gym-MujoCo Hopper Task Dataset}}
\label{fig:varying_actions_across_policies}
\end{figure}

\begin{figure}
  \includegraphics[width=0.45\textwidth, height=3cm]{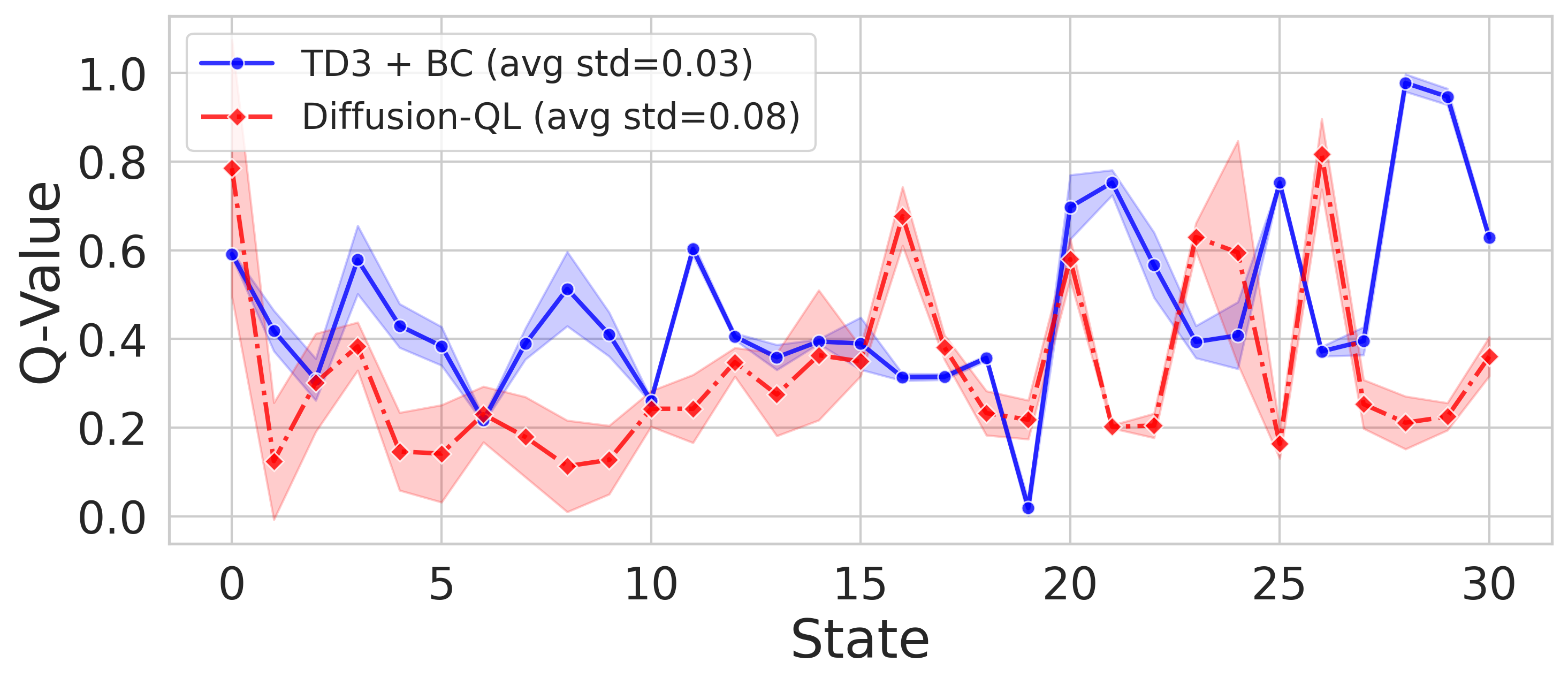}
  \caption{\textmd{Narrow spread of Q-values against contrastive actions in SOTA models}}
  \label{fig_generalization_behaviour}
  \vspace{-0.2in}
\end{figure}

\textbf{2. Normalized State Representation for Q-Value Estimation:} Figure \ref{fig:varying_actions_across_policies} indicates that counterfactual actions (actions taken by the alternate policy) are available. However, the state representations being different across the various policies, these actions would be treated as \textbf{out of distribution samples (OOD)} ~\cite{cql,iql,TD3,dr3,minimalist,diffusion}. Figure \ref{fig_generalization_behaviour} illustrates spread of Q-values against the actual and counterfactual actions (alternate policy) for a player across states. The exercise is repeated over two competitive SOTA models: TD3+BC \cite{minimalist} and Diffusion-QL \cite{diffusion}. We observe that the optimal policies learned by each of them are invariant to the counterfactual actions resulting in a narrow band of estimated Q-values. 
This is further exacerbated by the policy-specific bias towards time-varying confounders and covariates in the data, leading to varying actions (Further discussed in the Section \ref{sec:architecture}). This bias creates a need for a nuanced approach towards robust generalization of state representations across policies as previously highlighted in ~\cite{minimalist,fujimoto2019}. This bias is well controlled in the Gym-MuJoCo tasks as the states are fully observable.


\begin{figure}
\begin{minipage}{0.65\linewidth}
  \includegraphics[width=\linewidth, height=3cm]{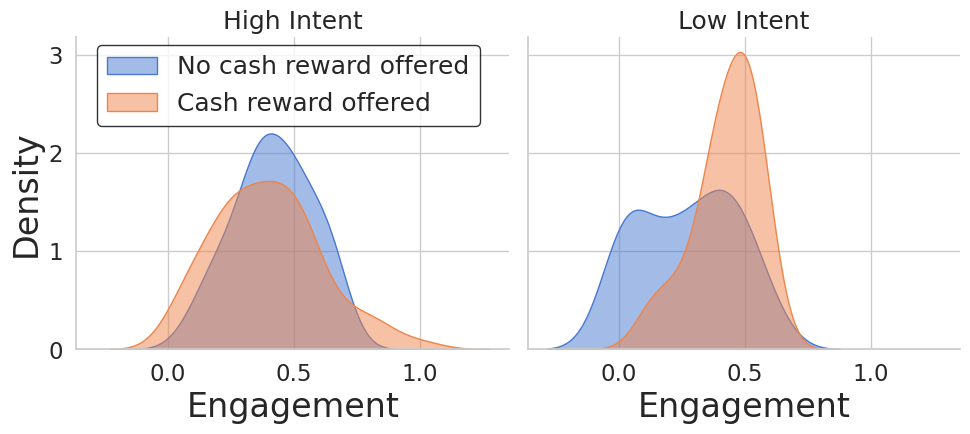}
  \caption{\textmd{Cash rewards do not add benefits for players with organic intent to play}}
  \label{fig_revenue_vs_organic_intent}
  \vspace{-0.2in}
\end{minipage}
\hfill
\begin{minipage}{0.33\linewidth}
\includegraphics[width=\linewidth]{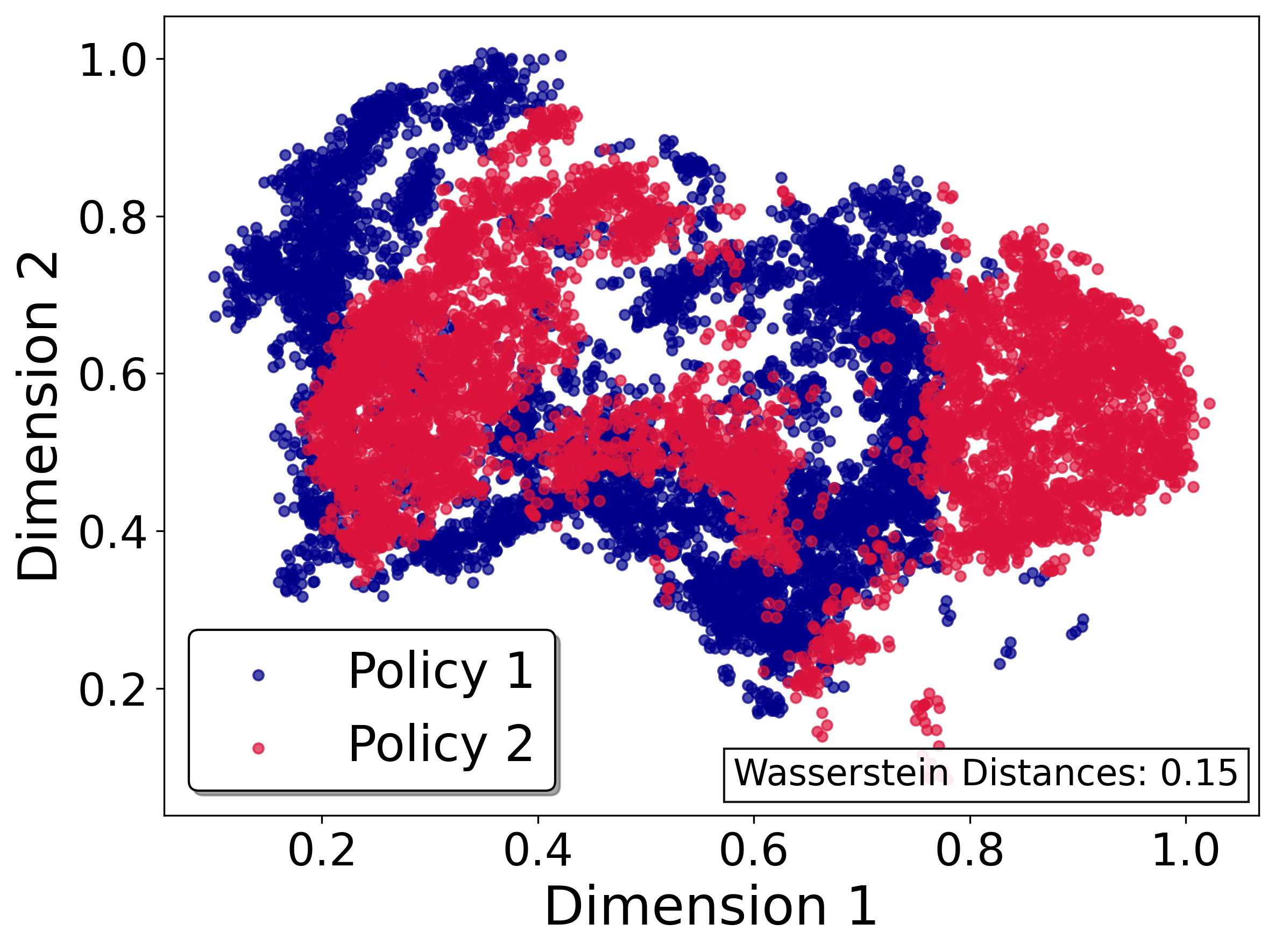}
\caption{\textmd{Poor generalization of the state space across policies in SOTA}}
\label{fig_state_isolation_reduction_diffusion}
\end{minipage}
\hfill
\end{figure}
\textbf{3. Engagement Volatility, Lack of Explainability:} Online gaming platforms experience significant fluctuations in player engagement (see Appendix~\ref{app:user_volatility}). In such businesses where short term revenue is actively tracked, a recommendation system focused on retention should do so with an explanation. The SOTA focuses on maximizing the long term reward with a limited focus on an instantaneous reward as a scalar quantity. Figure \ref{fig_revenue_vs_organic_intent} depicts that players with \textit{high organic intent} to play~\cite{cognitionnet,ARGO} may respond differently to cash rewards than those with low intent. This necessitates variable objective prioritization for a player over time with a specific focus on retention.

\textbf{4. Fast-Track Offline Counterfactual Exploration:} Figure \ref{fig_state_isolation_reduction_diffusion} illustrates state representation by Diffusion-QL~\cite{diffusion} model on Players Data. We observe a lower state isolation as compared to that seen in Figure \ref{fig_state_isolation_across_policy_classes}, but still persists to a large extent necessitating a further \textbf{online exploration} towards optimal policy learning which would incur considerable time given platform's volatility. This emphasizes on a need to better exploit the existing data in terms of counterfactual actions without risking Q-value overestimation.


\textbf{FAST-Q}, to the best of our knowledge is the first of its kind in addressing the above shortcomings while making the following \textbf{contributions}:

\textbf{1. Enable learning of balanced state representations:} We propose that, to reliably predict counterfactual outcomes it is essential to regularize the policy-specific bias between the players' state and action. FAST-Q proposes the use of \textit{Grandient Reversal Layer} (GRL) to learn \textbf{Balanced Representation} (BR) of the player’s states across various policies, an approach which has been long proposed ~\cite{ganin} but never leveraged in this context.

\textbf{2. Facilitate offline exploration of counterfactual actions:} FAST-Q relaxes the need for online exploration and brings down time to learn optimal policy by simply enabling exploration in tandem with offline data exploitation via counterfactual Q value estimates through the BR layer.

\textbf{3. Explainable objective prioritization:} There has not been much focus on explaining the objective prioritization of the recommendations. Some of the SOTA works pre-assign ~\cite{s-network} weights by empirical evaluations or by training separate Q-networks for each of the objectives contributing to the model complexity and its trainable parameters. FAST-Q proposes \textbf{decomposition of the Q-Values} via a \textit{complementary} loss which not just aids explainability but also keeps the Q-values bounded.

\textbf{4. Real-world evaluation on high-volatility gaming platform:} FAST-Q demonstrates \textit{at least} \textbf{0.15\%} increase in player returns, \textbf{2\%} improvement in lifetime value (LTV), \textbf{0.4\%} enhancement in the recommendation driven engagement, \textbf{2\%} improvement in the players' platform dwell time and an impressive \textbf{10\%} reduction in the costs associated with the recommendation, when compared with the most recent and notable SOTA methods on our volatile gaming platform.

\textbf{FAST-Q} proposes a new benchmark to facilitate a more accurate, explainable and fast-track exploration framework in offline RL. Our implementation is available at ~\cite{FAST-Q_git}

\section{Related Work}
\textbf{Policy Regularization:} Most prior methods for policy regularization in the context of offline RL rely on behavior cloning: BCQ \cite{fujimoto2019} constructs the policy as a learnable and maximum-value-constrained deviation from a separately learned Conditional-VAE ~\cite{cvaeRLNIPS} behavior-cloning model; BEAR ~\cite{BEAR} adds a weighted behavior-cloning loss via minimizing maximum mean discrepancy into the policy improvement step; TD3+BC ~\cite{minimalist}, our base model, applies the same trick as BEAR via maximum likelihood estimation (MLE); IQL~ \cite{iql} generalizes capacity of the function approximator to estimate best available action while balancing two conflicting objectives - behavior policy cloning and learning optimal policy; Diffusion-QL ~\cite{diffusion} builds on this ideology and proposes a diffusion model which is a conditional model with states as conditions and actions as outputs. 

\textbf{Multi-objective RL:} Most of the existing recommender systems try to balance the instant metrics and factors, focusing on improving the users’ implicit feed-back clicks ~\cite{wsdm,cikm}, explicit ratings ~\cite{www_multiobj,nips_multiobj}.  Work proposed in \cite{s-network} looks at user stickiness (LTV) by estimating multiple long term metrics via a separate network. However to compose the present reward from these objectives, it computes a weighted sum of these, where the weights were empirically set. Few adopt a multi-policy approach and learn a set of policies to obtain the approximate Pareto frontier of optimal solutions. \cite{multi-2} performs multiple runs of a single-policy method over different preferences. \cite{multi-1} simultaneously learn the optimal manifold over a set of preferences. \cite{multi-3} employs an extended version of the Bellman equation and maintain the convex hull of the discrete Pareto frontier. Most of these involve simultaneously training multiple networks increasing the overall complexity. Finally, a latest work proposed in ~\cite{nips-morl} adapts to multiple competing and changing objectives whose relative importance is unknown to the agent. In our case, however, the objectives are fixed.

\textbf{Counterfactual Learning:} \cite{ganin} proposed an approach inspired by the theory on domain adaptation suggesting that, for effective domain transfer to be achieved, predictions must be made based on features that cannot discriminate between the training (source) and test (target) domains. This approach promotes the emergence of features that are (i) discriminative for the main learning task on the source domain and (ii) indiscriminate with respect to the shift between the domains. This approach was later adopted by \cite{sebag,soleimani,ben-david}, all which have been well acknowledged. One such approach \cite{counterfactual} has applied this proposal in the context of predicting counterfactual outcomes for medical treatments. We adopt the basic premise and learning from all of these SOTA works in the context of building a generalized state representation.
\vspace{-0.05in}
\section{FAST-Q Architecture}
\subsection{Terminology and Problem Formulation}
We design the problem as a Markov Decision Process \textit{(MDP)} defined by $\langle S, A, R, \rho, \gamma \rangle$, with state space $S$, action space $A$, scalar reward function $R$, transition dynamics $\rho$, and discount factor $\gamma$ ~\cite{suttonandbarto}. The behavior of an RL agent is determined by a policy $\pi$ which maps states to a probability distribution over actions (stochastic policy).

\begin{figure*}
  \includegraphics[width=0.95\textwidth, height=5cm]{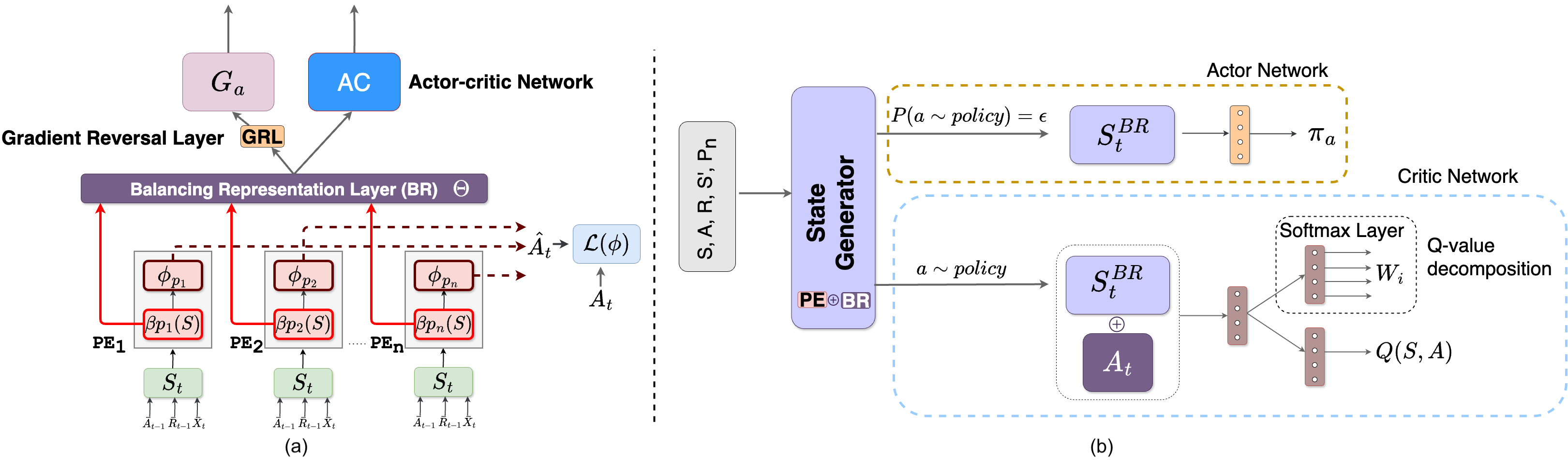}
  \caption{\textmd{FAST-Q Architecture. (a) Representing the PEs and their respective state representations feeding into the Policy Classifier and Actor-Critic (AC) Network, (b) Highlighting the flow of BRs through the AC Networks with Actor using $\epsilon$ factor to explore counterfactual and exploit exiting actions.}}
  \label{model_architecture}
\end{figure*}

\textbf{Data:} Consider a static data buffer B consisting of past data against \textit{n} different policies such that $B$ = $\{ B_{1} \cup B_{2} \cup B_{3} \dots B_{n} \}$. $B_{i} = \{(X_{1} , A_{0}, R_{0})_{i} \dots (X_{t} , A_{t-1}, R_{t-1})_{i}\}$ contains a challenge interaction sequences for players for policy $i$. Here, $X_{t}$ represents player gameplay features (see Table \ref{tab_game_features}), $A_{t}$ is the challenge served, and $R_t$ is the reward at time $t$.

\textbf{State:} The state at time step $t$, $S_t$ consists of the current game-play features of the user, the action performed, and the reward achieved to arrive at the current state, along with the state history. The history helps the MDP to capture the long term dynamics of the player. At the beginning, $S_{0}$ = $\{X_{0}\}$  which just contains user’s information, since there is no game play yet. At time step $t$, $S_{t}$ = $\{ X_{t} , A_{t-1}, R_{t-1}\}$ $\oplus \{S_{t-1}\}$. 

\textbf{Action:} A policy interacts with a user $u~\in~\mathcal{U}$ at discrete time steps, where $\mathcal{U}$ represents the universal set of users on our platform. At each time step $t$, it generates an action, $A_{t}~\epsilon~\mathcal{R}^{M}$. In our case, the action is 3-dimensional $(M=3)$: with the dimensions representing the number of games, the target score, and the cash reward upon completing the challenge. 

\textbf{Reward:} The reward $R_t \in \mathcal{R}^\mathcal{C}$ is tripartite ($\mathcal{C}=3$) at each time-step $t$, consisting of - \textbf{1) Dwell Time} is the time spent on the platform by the player while playing towards the recommended challenge; \textbf{2) Engagement} measures the time in days taken ($b$) by the player to return to the platform after \textit{seeing} the challenge, and is normalized as $\frac{6 - b}{6}$; and \textbf{3) Return Time} measures the time in days taken ($a$) by the player to return to the platform after \textit{completing} the challenge, and is normalized as $1- \sin{\frac{6 - a}{2}}$. Engagement measures the attractiveness of the challenge while the return time measures the game-outcome and the player experience from it.

\begin{table}[t]
  \footnotesize
  \begin{tabular}{ | c | l | }
\hline
\textbf{Dimension} & \textbf{Feature} \\
\hline
\multirow {3}{*}{Pre-Game} & Type of Rummy Format to play  \\
& Entry fee paid while entering the table\\
& CognitionNet \cite{cognitionnet} game behaviors,  predicted intent score \cite{ARGO}\\
\hline
\multirow {3}{*}{In-Game Choices} & Play slower than usual for focus \\
& Invalid declaration of the win \\
& Continue to play on bad cards (drop adherence~\cite{kddgames,pakddgames})\\
\hline
\multirow {3}{*}{Outcome} & Won vs. lost the game \\
& Amount of money actually lost\\
\hline
\end{tabular}
  \caption{\textmd{Sample Game Features}}
  \vspace{-0.3in}
  \label{tab_game_features}
\end{table}

Our existing challenge recommendation system periodically shuffles assignment of players among policies every $x$ days, with $x$ ranging from 14, 21, 30 up to 45 days. Since we deal with a continuous action prediction problem we go with a Actor-Critic Architecture.


\subsection{FAST-Q Network}
 Estimation of Q-value for a counterfactual action on a state history from a different policy is not trivial primarily due to 1) the auto-regressive nature of the critic which is influenced by the past history over $T$ steps 2) high state disparity across policies leading to a poor generalizations across unobserved states as seen in Figure \ref{fig_state_isolation_across_policy_classes}. Obtaining a \textit{Balancing-Representation~(BR)} on the state representation becomes crucial. Once we solve this problem, we can now enable exploration on actions never exposed to a user being treated by a fixed policy for $x$ days. Figure \ref{model_architecture} shows architecture and loss function flow in FAST-Q~\footnote{See Appendix~\ref{app:algorithm} for the FAST-Q algorithm}. 
 
\subsection{Policy Expert (PE)} Each Policy has its own interpretation of the state of the user, based on which a new action is recommended. We propose Policy Experts layer as the first component of FAST-Q.  Each PE consists of a base LSTM network \cite{lstm}, followed by a dense layer. The LSTM first builds its own hidden state representation, $\beta(S_{t})$ using the history of actions $\bar{A}_{t-1}$, rewards $\bar{R}_{t-1}$ and player's time-variant game-play features $\bar{X}_{t-1}$, along with the present game features $X_{t}$. The dense layer $\phi$ is further applied to predict an action $\hat{A_{t}}$ over this representation. PEs are trained on the weighted MSE loss of the three predicted action dimensions w.r.t the actual values seen in the static data.
$\mathcal{L}(\phi)$ = $ \parallel A_{t} - \phi(\beta(S_{t})) \parallel_{2} $, where $\phi(\beta(S_{t}))$ is represented as ~$\hat{A_{t}}$ as in Figure \ref{model_architecture}(a)
\subsection{Counterfactual Network} 
\label{sec:architecture}
State-action pairs can be used to train a supervised critic model to predict: $\mathrm{E}[R_{t} | S_{t}, A_{t}]$. However, without adjusting for the bias introduced by time-varying confounders, this model cannot be reliably used for making causal predictions, specifically pertinent when switching policies \cite{Robins2008,schulamSaria}. The counterfactual network removes this old policy bias through domain-adversarial training and estimates the counterfactual outcomes as $\mathrm{E}[R_{t} | \Theta(\beta(S_{t})), A_{t}]$, for the balanced state representation $\Theta(\beta(S_{t}))$. 

\textbf{Balancing Representation (BR):} The state $S_{t}$ contains the time-varying confounders - $\bar{X}_{t},~\bar{A}_{t-1}, ~\bar{R}_{t-1}$, which biases the treatment assignment $\hat{A_{t}}$ coming from $\phi(\beta(S_{t}))$. The representation $\Theta(\beta(S_t))$ is built by applying a fully connected dense layer to the hidden state representation $(\beta(S_t)$ of the PE. $\Theta$ is the representation function that maps the action-specific state representation coming from the individual PEs to an unbiased action invariant representations such that:
\newline
$\mathrm{P}[\Theta(\beta_{p1}(S_t))~|~ \phi_{p1}(\beta_{p1}(S_t))] =~ \dots ~ = \mathrm{P}[\Theta(\beta_{pn}(S_t))~|~ \phi_{pn}(\beta_{pn}(S_t))]$, where $\beta{p_n}$ represents the hidden state from $n^{th}$ PE as seen in the Figure \ref{model_architecture} (a). To achieve this, we integrate the domain adversarial training framework proposed by ~\cite{ganin} and extended by ~\cite{sebag, counterfactual} to a multi-domain learning setting. In our case, the different actions at each time step are from different policies.

\begin{figure}
\includegraphics[width=0.3\textwidth, height=6.5cm]{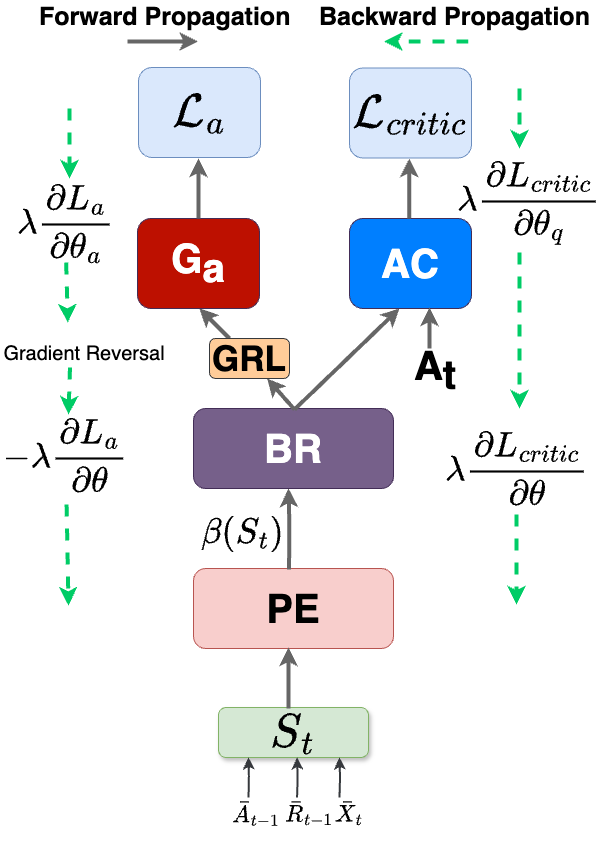}
\caption{\textmd{Training procedure for building balancing representation}}
\label{br_process}
\end{figure}

\textbf{Training Adversarially Balanced Representations:} One can construct a policy-invariant representation of the state by minimizing the distance between representations of the same state coming from different PEs. As suggested in \cite{ganin} we use the concept of minimizing the divergence between these representations by proposing an adversarial framework for domain (policy) adaptation by building a representation which achieves maximum error on a policy classifier and minimum error on an outcome predictor (critic loss in our case) as illustrated in the Figure \ref{br_process}.


Let $G_{a}(\Theta(\beta(S_{t})); \theta_{a})$ be a policy classifier with parameter $\theta_{a}$. Then the policy classification loss $\mathcal{L}_{a}$ is defined as:

$\mathcal{L}_{a} = -\sum_{p=1}^{n} \mathbb{1}[P_{t} = p].\log(G_{a}(\Theta(\beta(S_t));\theta_{a}))$

where $P_{t}$ is the policy at time t for a user and $\mathbb{1}[.]$ is the indicator function. The output layer of $G_{a}$ uses softmax activation. To train the counterfactual network using backpropagation, we use the \textbf{Gradient Reversal Layer} (GRL)~\cite{ganin}. To maximize the entropy loss, we sample data inversely proportional to the predicted policy class distribution in the previous iteration. This gave us the best results. 
\subsection{Critic Network} 
We follow the traditional method of Actor-Critic Training. The optimal $Q^*(S_t, A_t)$, having the maximum expected discounted reward achievable by the optimal policy, should follow the optimal Bellman equation~\cite{suttonandbarto}: 
$Q^{*}(S_t, A_t) = \mathbb{E}_{S_{t+1}} [ R_{t} + \gamma~\underset{A_{t+1}}{max}~Q^{*}( S_{t+1}, A_{t+1} | S_{t}, A_{t})]$
In FAST-Q, $S_t$ is transformed to $S_t^{BR}$, where $S^{BR}_t=\Theta(\beta(S_t))$. Henceforth, $S_t^{BR}$ will be used as the state representation. Critic network with parameter $\theta_{q}$ is used to estimate Q-value whose loss is defined as: \\
~$\mathcal{L}_{\textmd{TD}}(\theta_{q})$ = $E_{(S_{t},A_{t},R_{t},S_{t+1}) ~\sim ~B} [ ( y_{t} - Q(S_t^{BR},A_{t} ;\theta_{q}))^{2} ]$\\
~~~where $y_{t}$ = $R_{t} + \gamma~\underset{A_{t+1}}{max}~Q\left(S_{t+1}^{BR}, A_{t+1}; \theta_{q} \right)$

\subsubsection{\textbf{Overestimation in critic networks:}} \cite{TD3} suggests building on Double Q-learning, by taking the minimum value between a pair of critics to limit Q-value overestimation. They also draw the connection between target networks and overestimation bias, and suggest delaying policy updates to reduce per-update error and further improve performance. We adopt both these proposals. 

\subsubsection{\textbf{Q-Value decomposition:}} Additionally, we task FAST-Q model with complimentary objectives of predicting the various reward metrics as a part of the critic's Q-value learning. Figure \ref{model_architecture}(b) depicts that the penultimate layer of the critic is competitively trained to produce two alternate layers. Dense layer is applied on one to produce Q-value and a softmax layer is applied on the other to derive the 4 weights. Weights $w_{1}$, $w_{2}$, $w_{3}$ are trained to estimate the following objective metrics:
$$
\mathcal{L}_{\textmd{decomp}} = \frac {\sum_{i=1}^{C} [(R_{t}^{i} - w_{i}~ *~ Q(S_{t}^{BR},A_{t}))^2]}{C}
$$

where C is the number of reward dimensions. The fourth weight is primarily used as an \textit{overflow} weight, to stabilize the training and allow for any uncounted objective affecting the Q-value. Finally, cost of the recommendation is not controlled due to fairness and ethical reasons and is always a factor of the challenges shown to the player.
The final Critic loss, with $\alpha$ derived experimentally to 0.75 is represented as: 
$$
\mathcal{L}_{\textmd{critic}}= \alpha~\mathcal{L}_{\textmd{TD}}(\theta_{q}) + (1 - \alpha)~\mathcal{L}_{\textmd{decomp}}
$$

\subsubsection{\textbf{Stepwise change of the discount factor:}} 
\begin{figure}
  \includegraphics[width=0.45\textwidth, height=3cm]{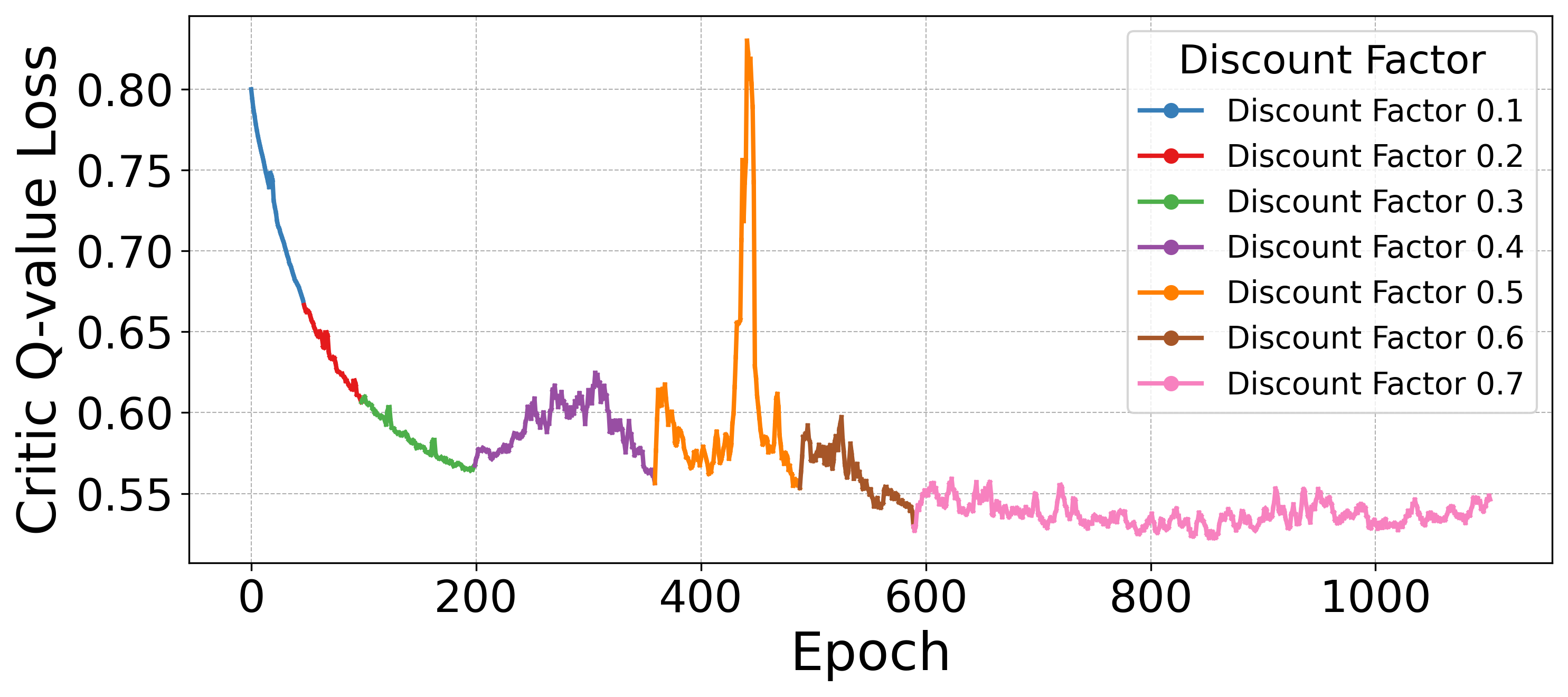}
  \caption{\textmd{Training and loss convergence in FAST-Q with increamental $\gamma$ ~lift}}
  \label{fig_FAST-Q_training_Convergence}
  \vspace{-0.2in}
\end{figure}

Since FAST-Q learns the individual rewards and associates MSE's on those as a part of the critic loss, discount factor $\gamma$ had to be gradually increased. Figure \ref{fig_FAST-Q_training_Convergence} shows the incremental training of FAST-Q with stepwise increase in the discount factor $\gamma$ from 0.1 until 0.7. The discount factor is only increased upon stabilization of the network losses, as seen in the Figure \ref{fig_FAST-Q_training_Convergence}. This was done to ensure its gradual convergence. One of the downsides of Q-value decomposition was that, we were not able to train FAST-Q beyond the $\gamma$ value of 0.7.

\subsection{Actor Network} Like the critic network, the actor network is also trained on $S_{t}^{BR}$ for the current policy. The actor predicts an action, for which the critic's Q-value is maximized. We adopt the strategy suggested in ~\cite{minimalist} and add a Behavior Cloning (BC) term to the policy update algorithm to handle OOD action sampling. The actor loss is defined as:
$$
\mathcal{L}_{\textmd{actor}} =  - \lambda~E[Q(S_{t}^{BR}, \pi(S_{t}^{BR}))] + (\pi(S_{t}^{BR}) - A_t)^2
$$
where $\lambda = \frac{\alpha}{\textmd{mean}(|Q|)}$ as defined in ~\cite{minimalist}, $A_t$ is the actual action under a policy. We use $\alpha=2.5$ in our experiments, as recommended in ~\cite{minimalist}.
Balancing the state representation enables counterfactual action evaluation. To facilitate this further, we propose Fast-track exploration. 
\subsubsection{\textbf{Fast-track exploration with counterfactual actions:}}
While training on a particular discount factor we gradually increase the exploration factor, $\epsilon$ of the actor from $0.1$ to $0.5$. During exploration, we randomly choose a counterfactual action. Hence, the second term in $\mathcal{L}_{\textmd{actor}}$ replaces $A_t$ with $\hat{A}_{cp}$, where $\hat{A}_{cp}=\phi_{cp}(\beta_{p_{cp}}(S_t))$ and cp indicates a counterfactual policy.

Finally, though in this experiment we have used a tripartite reward and tripartite action dimensions, one can re-configure these dimensions as per their applicability. 
\section{Evaluation}
\subsection{Experimental Setup on RummyCircle Platform}

We collected a large scale dataset including over 2 million recommendations using each of the 3 policies (as initially discussed in the section \ref{sec:intro}) for a period of 4 months. Policy-3 enabled FAST-Q to learn ``no challenge'' as also a viable action. 

We implemented FAST-Q using PyTorch \textit{1.13.1+cu117} with cuda and training has been done on NVIDIA A10G GPU on Databricks. See Appendix~\ref{app:hyperparameters} for more details on the hyperparameters.


\subsection{Offline Evaluation}
We consider three objectives - 1) Dwell Time, 2) Engagement and 3) Return Time. The trained FAST-Q model was retrospectively run on a held-out dataset to derive the following insights:



\subsubsection{Objective Prioritization:} 
\label{sec:offline_obj}
\begin{figure}
  \includegraphics[width=0.42\textwidth, height=3cm]{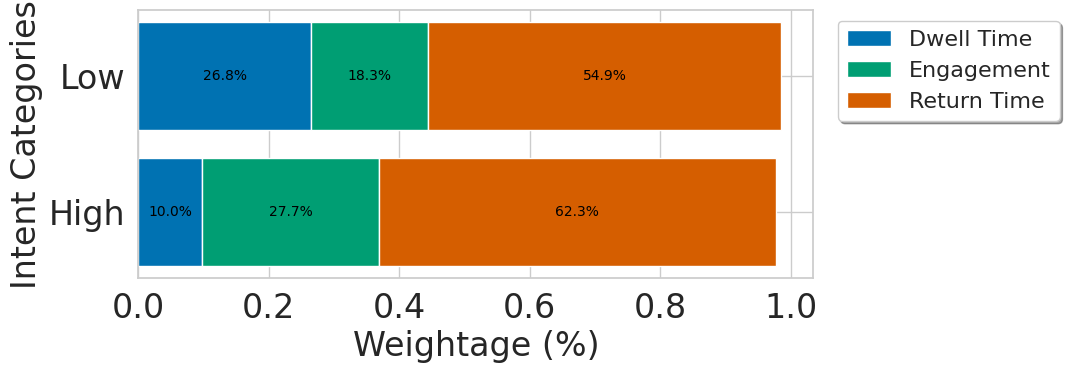}
  \caption{\textmd{Varying objective focus deployed by FAST-Q based on Players' intent feature.}}
  \label{fig_intent_objective_pie}
\end{figure}

Figure \ref{fig_intent_objective_pie} shows normalized distribution of the objectives for players with high organic intent~\footnote{intent values are available to us as a part of the game-play features~\cite{ARGO}} to play, FAST-Q puts major focus on the return time followed by players' engagement to keep them active on the platform. Whereas, when the intent is low the objective shifts more towards improving players' dwell time on the platform for the day with proportionate reduction in other objectives. 

\begin{figure}
  \includegraphics[width=0.45\textwidth, height=3cm]{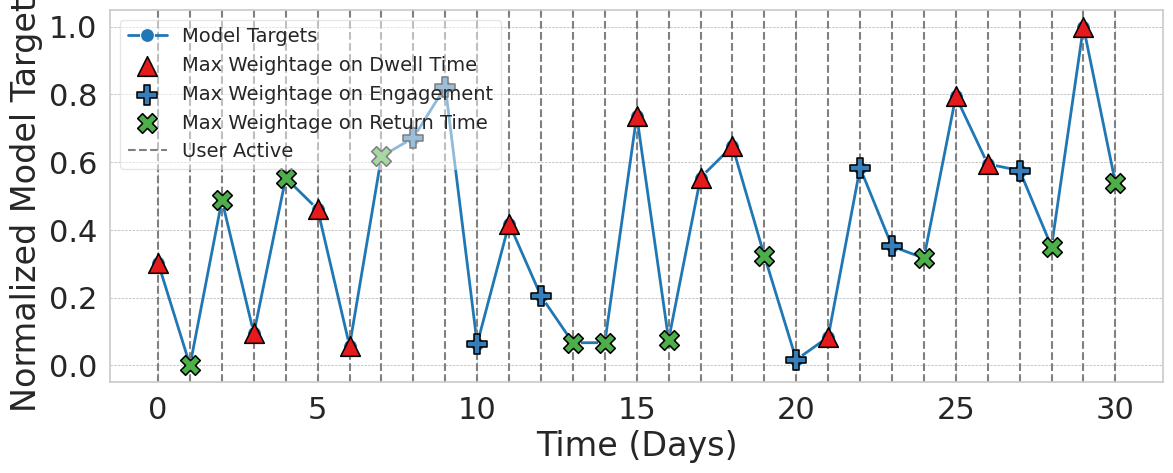}
  \caption{\textmd{Good balance of objectives by FAST-Q for an active user}}
  \label{fig_active_user_objective_focus}
  \vspace{-0.2in}
\end{figure}

\begin{figure}
  \includegraphics[width=0.45\textwidth, height=3cm]{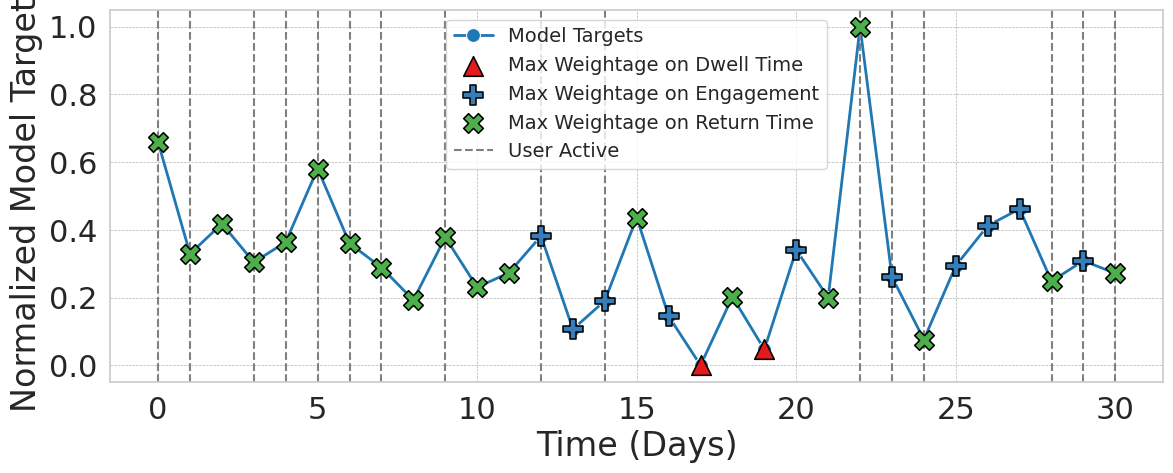}
  \caption{\textmd{FAST-Q primarily focuses on the return time for a seldom active user}}
  \label{fig_inactive_user_objective_focus}
  \vspace{-0.2in}
\end{figure}

Figure \ref{fig_active_user_objective_focus} shows that for an active player, FAST-Q shows good balance of objectives throughout the time under observation vs. a less active player in the Figure \ref{fig_inactive_user_objective_focus}, where it primarily favors return time with intermittent focus on engagement. Figure~\ref{fig_intent_objective_pie} depicts FAST-Q focus at a platform level while Figures~\ref{fig_active_user_objective_focus}, \ref{fig_inactive_user_objective_focus} depict treatments at a player level. 

\begin{figure}[h]
  \includegraphics[width=0.45\textwidth, height=4cm]{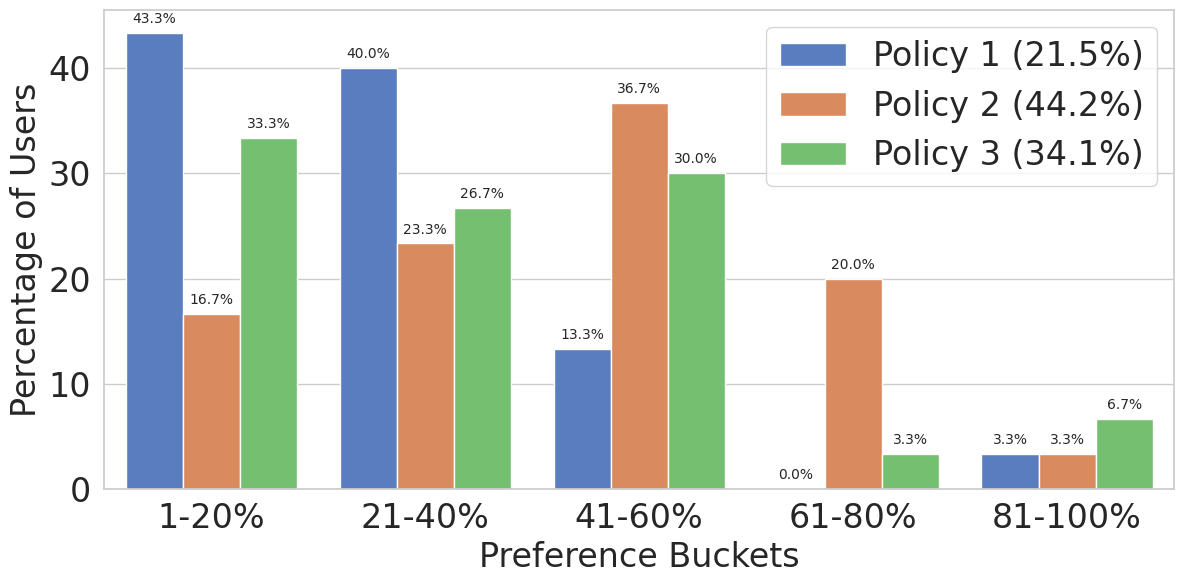}
  \caption{\textmd{Proportion of policy exposure on our platform after FAST-Q}}
  \label{fig_actual_vs_counterfactual_weightage}
\end{figure}

 Offline analysis on a select group of players revealed that a blend of actions from both policies could yield improved outcomes, as illustrated by the solid blue (policy switch) line connecting different policies over time (highlighting policy switches) in the Figure \ref{fig:varying_actions_across_policies}. This conclusion was derived by evaluating player engagement and revenue in statistically similar states where actions from different policies were applied. Prior to FAST-Q, Policy 1, Policy 2 and Policy 3 were split in a 40\%, 40\% and 20\% split over our platform players. Such that each player was served with the same policy exclusively. Figure \ref{fig_actual_vs_counterfactual_weightage} shows the new distribution of policy preference by FAST-Q. In conformance of the retrospective evaluation in the in Figure \ref{fig:varying_actions_across_policies}, we observe that FAST-Q favored the actions from Policy-1, Policy 2 and Policy 3,~ 21.5\%, ~44.2\%, and ~34.1\% of the times respectively, on an average. This validates our hypothesis substantiating a need to shuffle between the policies.

\subsubsection{Addressing State Space Sparsity and confounders in BR:} 
\begin{figure}
  \includegraphics[width=0.45\textwidth, height=3cm]{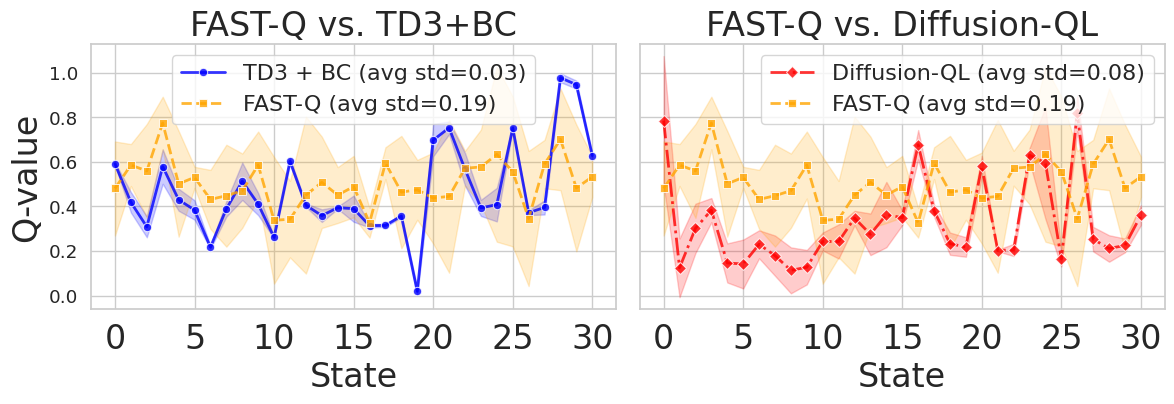}
  \caption{\textmd{Better spread of Q-values against the actual and counterfactual actions in FAST-Q vs SOTA}}
  \label{fig_generalization_behaviour_part2}
\end{figure}

\begin{wrapfigure}{l}{0.45\linewidth}
\includegraphics[width=\linewidth]{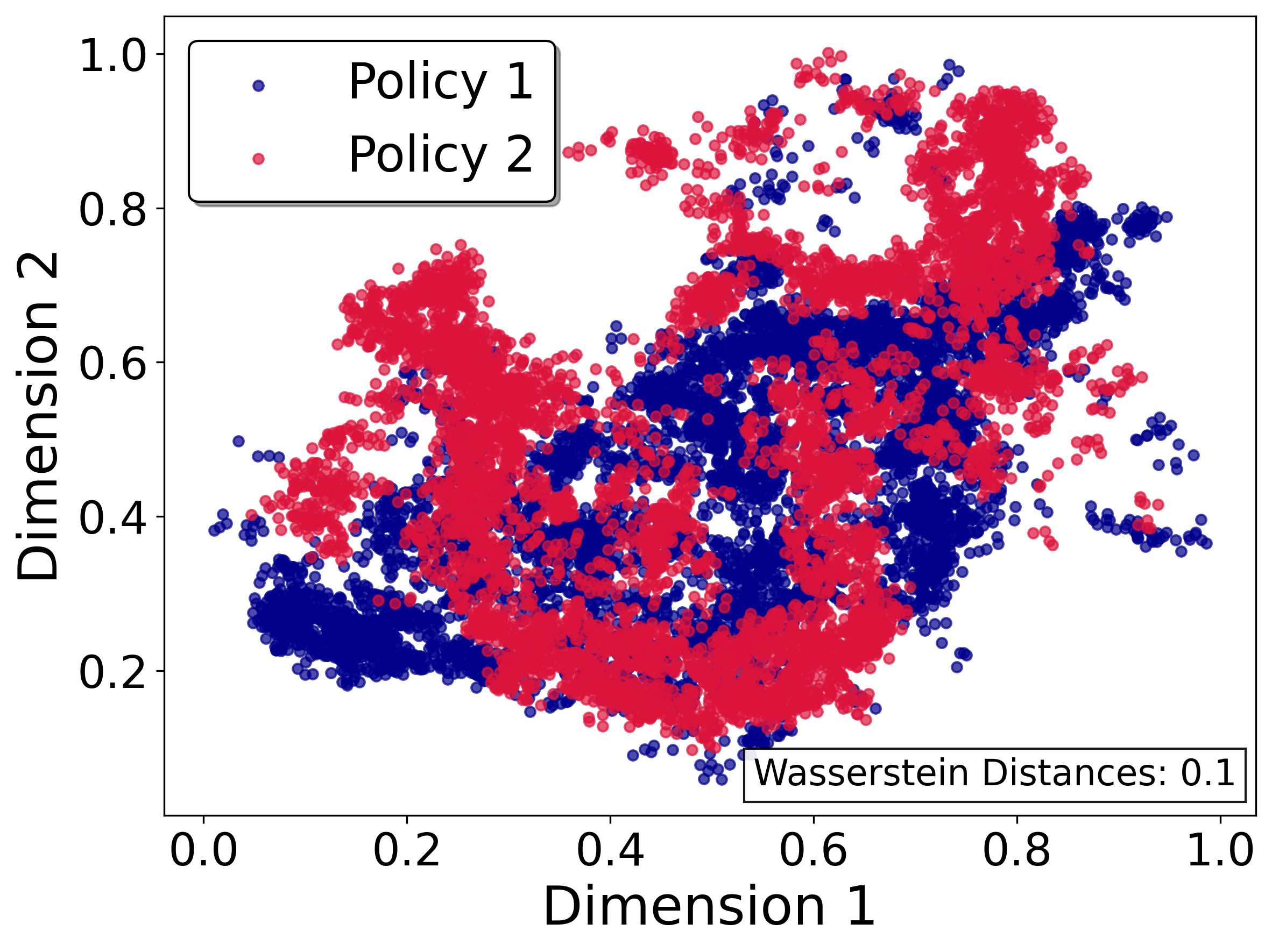}
\caption{\textmd{\small Well generalized (blended) state representations after BR}}
\label{fig_BR_with_FAST-Q}
\hspace{-0.45 in}
\vspace{-0.25 in}
\end{wrapfigure}

Figure \ref{fig_BR_with_FAST-Q} illustrates much balanced state representation across the policies compared to that seen in Figures \ref{fig_state_isolation_across_policy_classes} and \ref{fig_state_isolation_reduction_diffusion}. Figure \ref{fig_generalization_behaviour_part2} shows the spread of Q-values against the actual and counterfactual
actions (alternate policy) for a player across states with FAST-Q against SOTA as previously discussed in the Section \ref{sec:intro}. FAST-Q leverages BR for predictions and this results 2.3x and 6.3x higher spread of Q value estimates against Diffusion-QL and TD3+BC respectively. 

\begin{figure}
  \includegraphics[width=0.45\textwidth, height=3cm]{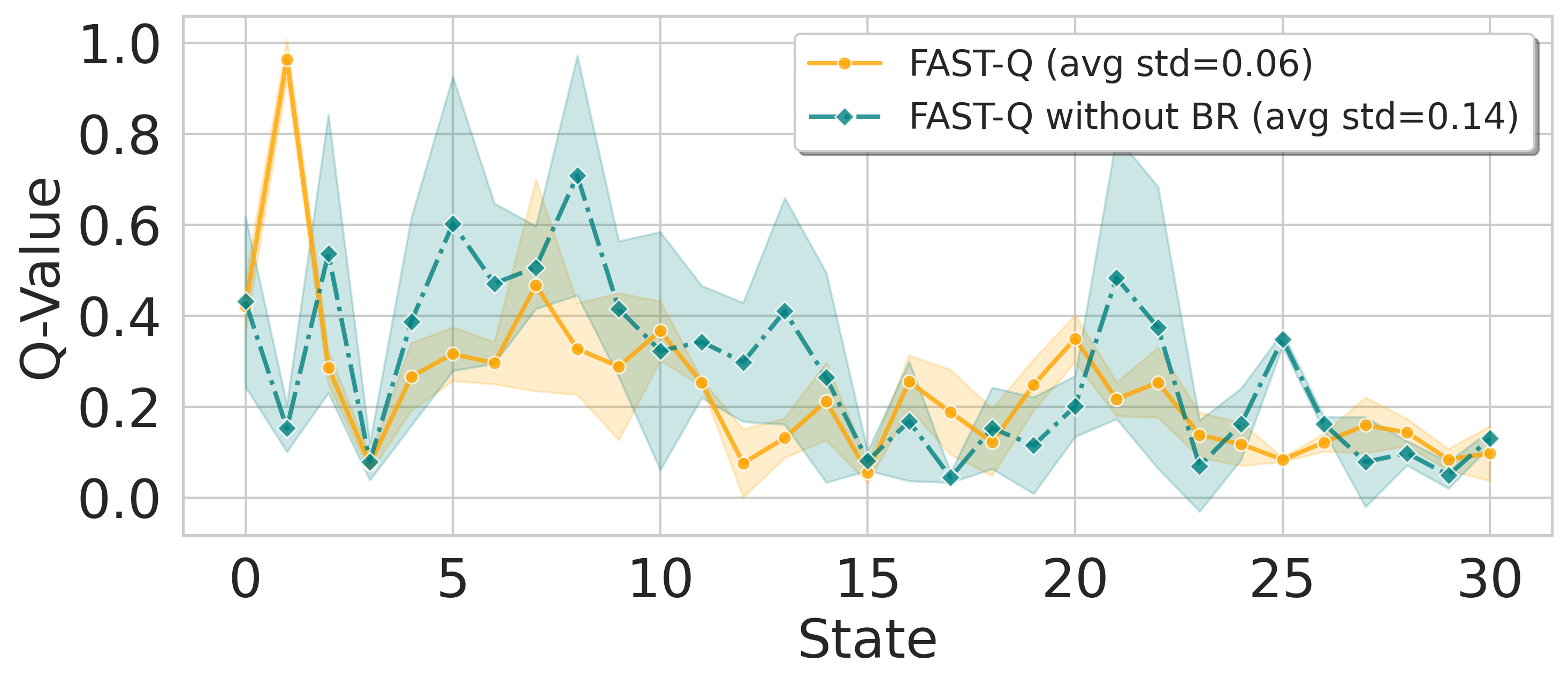}
  \caption{\textmd{Uncertainties on Q-value estimation on a state is 2.3x higher without the BR layer}}
  \label{fig_uncertainityy_estimates}
\end{figure}

To further assess the confidence of FAST-Q in predicting the Q-values, we added variational 
dropouts \cite{variational_dropouts} in the PE networks and trained FAST-Q with and without the BR loss. 
Figure \ref{fig_uncertainityy_estimates} shows that FAST-Q without the BR loss shows 2.3x higher variation in the Q-value estimates against multiple predictions on a state representation. This is because the state representation without the BR training is overly discriminative of the policy under evaluation such that slight perturbation in the state representation due to regularization (variational dropouts) changes its interpretation and hence the corresponding Q-value. 

\subsubsection{Time Speedup by Exploration:} 
\begin{figure}
  \includegraphics[width=0.45\textwidth, height=3cm]{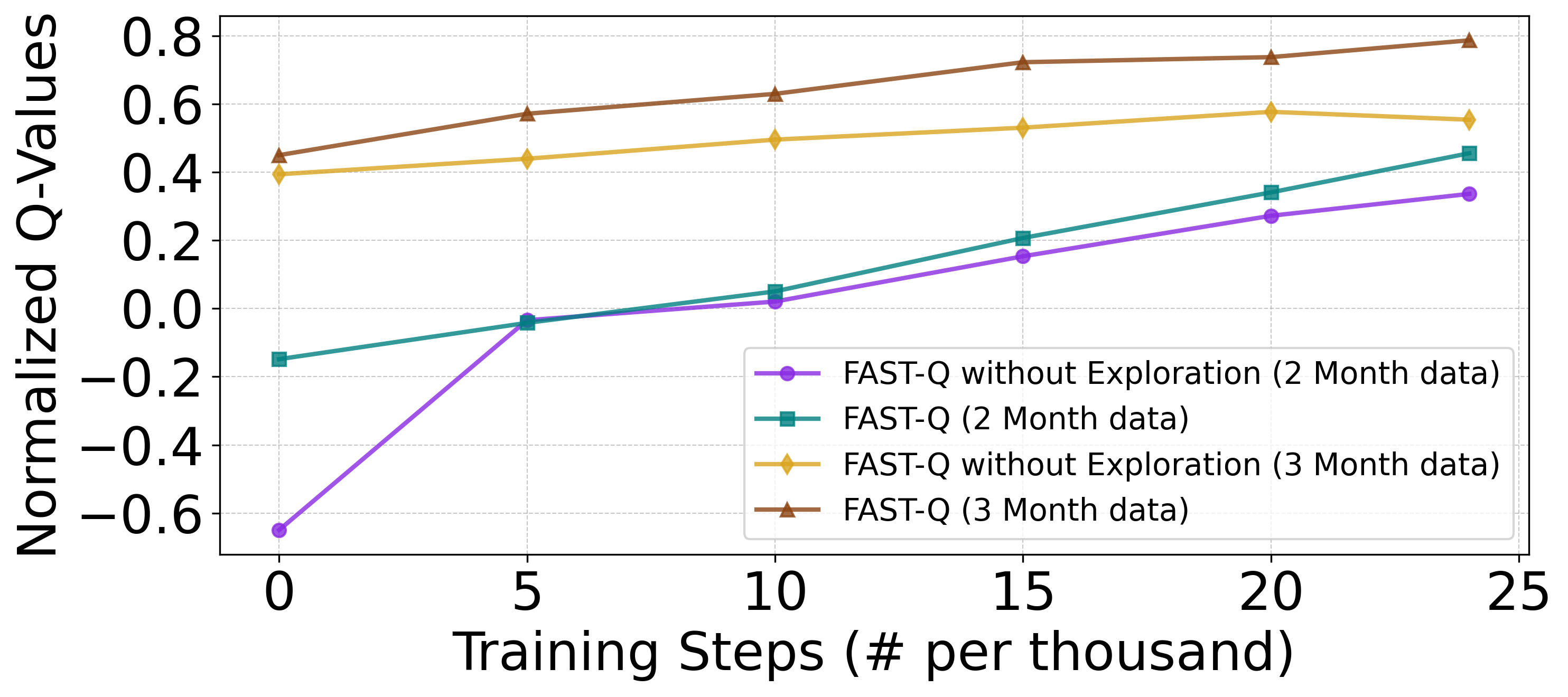}
  \caption{\textmd{Time speed up with Offline Exploration on Counterfactuals.}}
  \label{fig_Q_values_SOTA_offline}
\end{figure}

Figure \ref{fig_Q_values_SOTA_offline} shows a comparative analysis of the model's performance, normalized with respect to the Q-values of the FAST-Q model trained on the complete dataset, highlighting the impact of exploration. The graph demonstrates that enabling the exploration component accelerates the learning process. For instance, the return for the model trained with exploration on 2 months of data is equivalent to model trained without exploration on 3 months of data.

\subsubsection{Ablation Comparison:}  
\begin{figure}
  \includegraphics[width=0.45\textwidth, height=3cm]{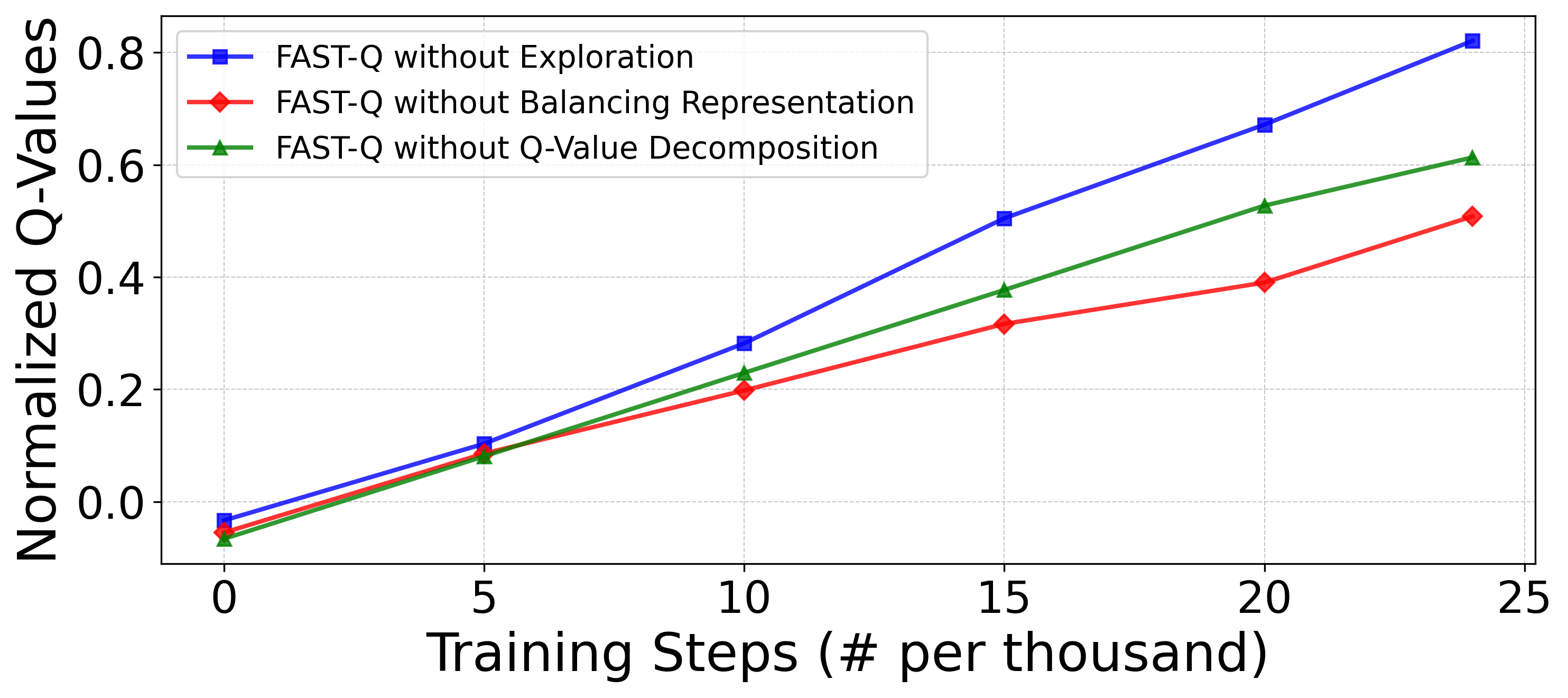}
  \caption{\textmd{Drop in Q-values w.r.t the deployed model with Ablation on the FAST-Q Network components}}
  \label{fig_ablation_Q}
\end{figure}

Figure \ref{fig_ablation_Q} shows comparative
analysis of the model’s performance on its ablated versions, normalized with respect to
the Q-values of the final FAST-Q model used in our online A/B testing path. We observe higher estimated Q-values with FAST-Q vs. its ablated versions. We see that only removing the BR layer shows a significant \textbf{50\%} drop in the Q-values. Only removing offline exploration drops the maximum achievable Q values by at least \textbf{18\%}. Finally we see an impressive \textbf{40\%} drop in the maximum achievable Q-Values by only turning off Q-value decomposition for multi objective prediction. These results confirm our understanding that multi objective learning along with providing meaningful explanations also helps in realistic and better estimation of the Q-values. 


\subsection{Online Experiment on RummyCircle Dataset}
To further assess the real-world performance, we conducted an online A/B test by assigning 20\% of users to our FAST-Q model and  20\% each to the two SOTA models TD3+BC~\cite{minimalist} and Diffusion-QL~\cite{diffusion} and 20\% each to the two in-house policies (1 \& 2) on the RummyCircle platform. Policy 3 was not on the test path in this experiment, however recollect that FAST-Q had used Policy 3 data during its offline training. We directly leveraged the published code and only modified data processing pipeline of TD3+BC ~\cite{minimalist_git} and Diffusion-QL ~\cite{Diffusion_git} to process our game play sequential data (sample data at~\cite{FAST-Q_git}).
This online experiment took place in the first week of October, spanning a 21-day period. Our model consistently outperformed during the 21 days, and it has been deployed for the full user base now (spanning over 120 million unique players). For A/B testing, the user population was fairly split based on their past activity and LTV on the platform. The results shown here have all been adjusted for the pre-bias.

\subsubsection{Recommendation Feeds:} The respective policy agent determines the tripartite action that offers the maximum Q-value based on the current state. This action is shown to the players as soon as they login into the RummyCircle online application. 

\begin{figure}[htp]
\subcaptionbox{(a) \% lift in Engagement and Return Time with FAST-Q w.r.t SOTA \label{fig_eng_ret}}{
  \includegraphics[width=0.45\textwidth, height=3cm]{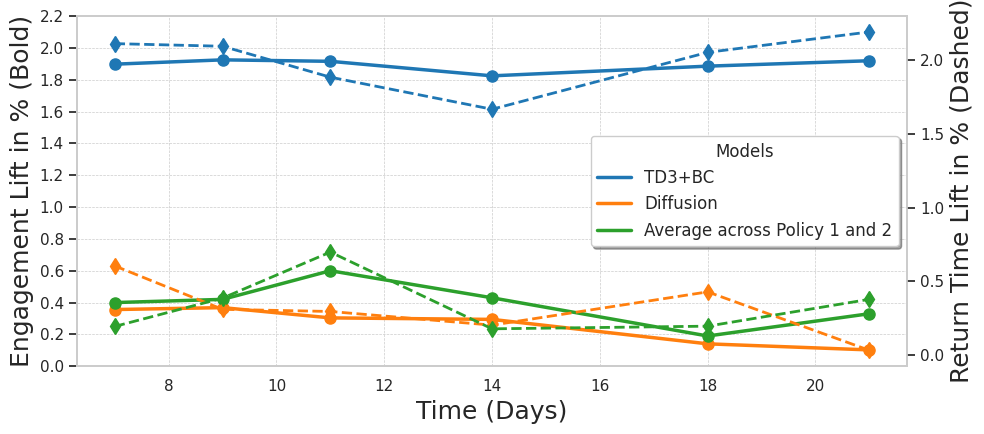}
}
\vspace{3ex}

\subcaptionbox{(b) \% lift in Dwell Time and drop in cost with FAST-Q w.r.t SOTA \label{fig_rev_cost}}{
  \includegraphics[width=0.45\textwidth, height=3cm]{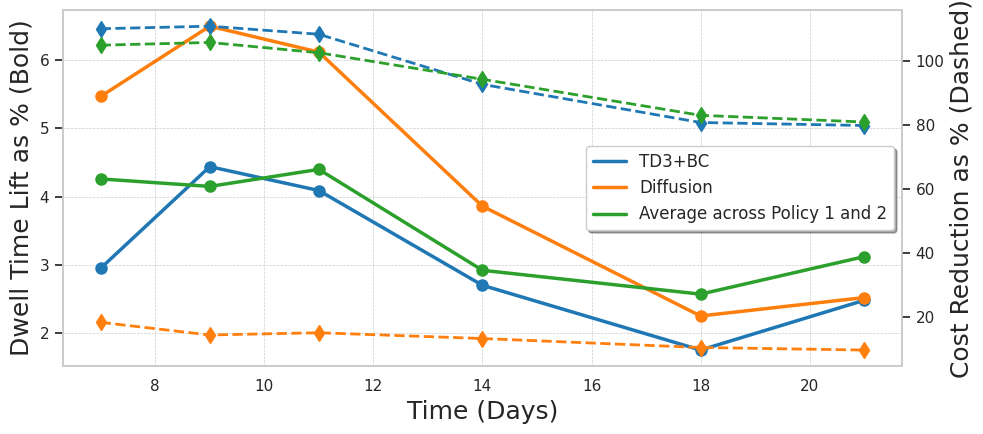}
}
\caption{\textmd{Changes in $\Delta$ Engagement, $\Delta$ Return Time, $\Delta$ Dwell Time, $\Delta$ Cost within 21 days}}
\label{fig_online_results}
\end{figure}

\subsubsection{Objective Metrics Performance:} To provide a more intuitive representation of the online results, we employed commonly used metrics: where $\Delta~\% = \left( \frac{\textmd{Metric}_{\textmd{FAST-Q}}}{\textmd{Metric}_{\textmd{SOTA}}} - 1 \right) \times 100$, except in the case of the cost metric where the positions in the formula are reversed. Figure \ref{fig_online_results}(a) shows \% lift in engagement and return time on the platform when challenges were served with SOTA recommendations vs. with FAST-Q. We see a minimum of 1.8\%, 0.15\% engagement lift and 1.6\% and 0.15\% return time lifts w.r.t TD3+BC and Diffusion-QL models respectively. Policy 1 and 2 performed identically and hence we report their combined results all through. With them, FAST-Q shows at least 0.2\% lift in both the engagement as well as return times. 
Figure \ref{fig_online_results}(b) shows \% lift in dwell time and  \% drop in the cost (indicating the reward offered and paid to the player). The $\Delta$'s for both metrics are throughout positive showing maximum lift in dwell time and maximum cost saving both simultaneously w.r.t the existing platform policies. FAST-Q also shows at least 2\% improvement in the dwell time w.r.t all the SOTA. In spite of retaining higher engagement and dwell time on a daily basis, FAST-Q also shows cost reduction of at-least 10\% w.r.t the Diffusion-QL model. FAST-Q shows an impressive reduction of cost by 80\% at D21 (21 days) w.r.t the exiting platform policies as well as the TD3+BC.


\subsubsection{Results with A/B Experiment Conclusion Framework:} 
\begin{figure}[htp]
    \centering
    \subcaptionbox{Old Policies \label{fig_AB_old_policy}}{
        \includegraphics[scale=0.118]{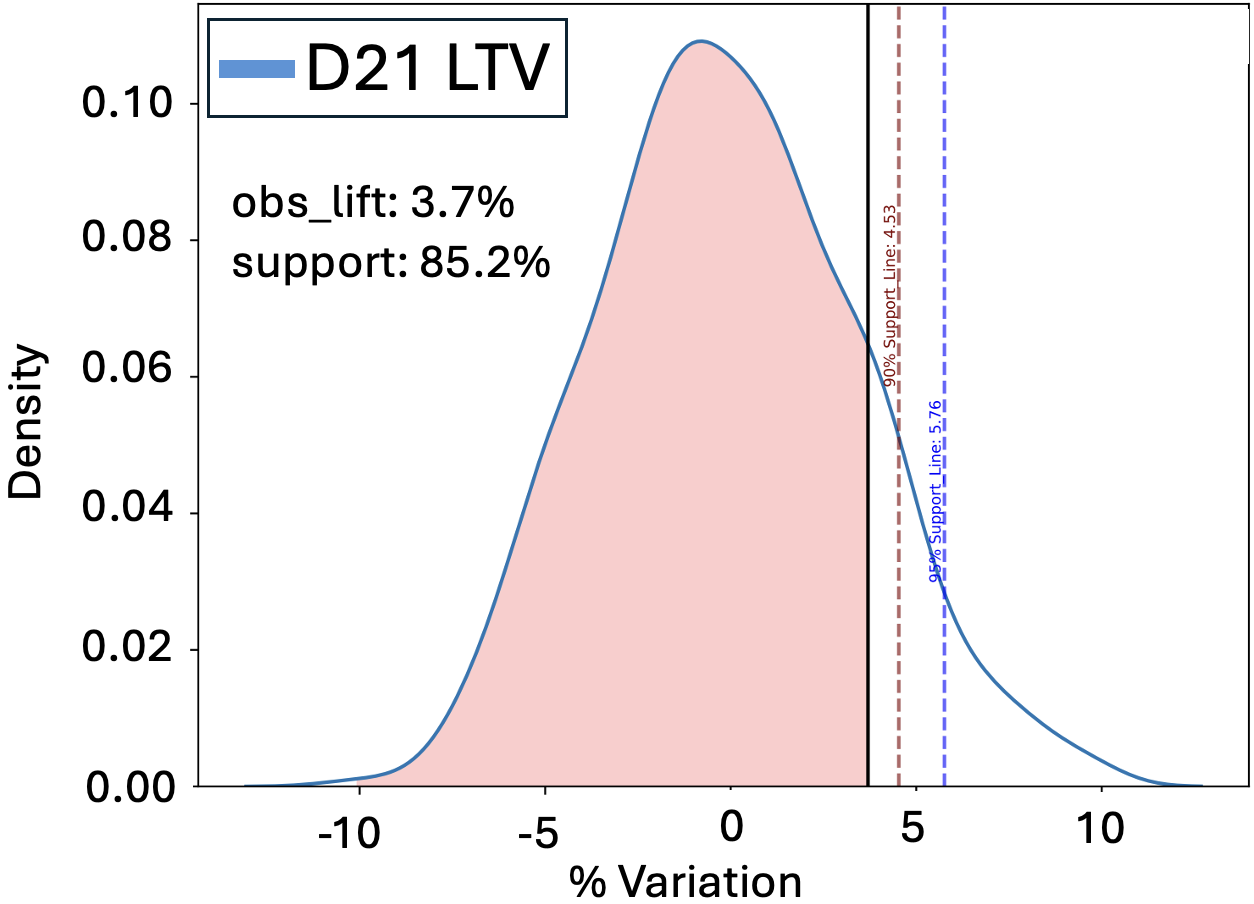}   
    }
    \subcaptionbox{TD3+BC\label{fig_AB_minimalist}}{
        \includegraphics[scale=0.118]{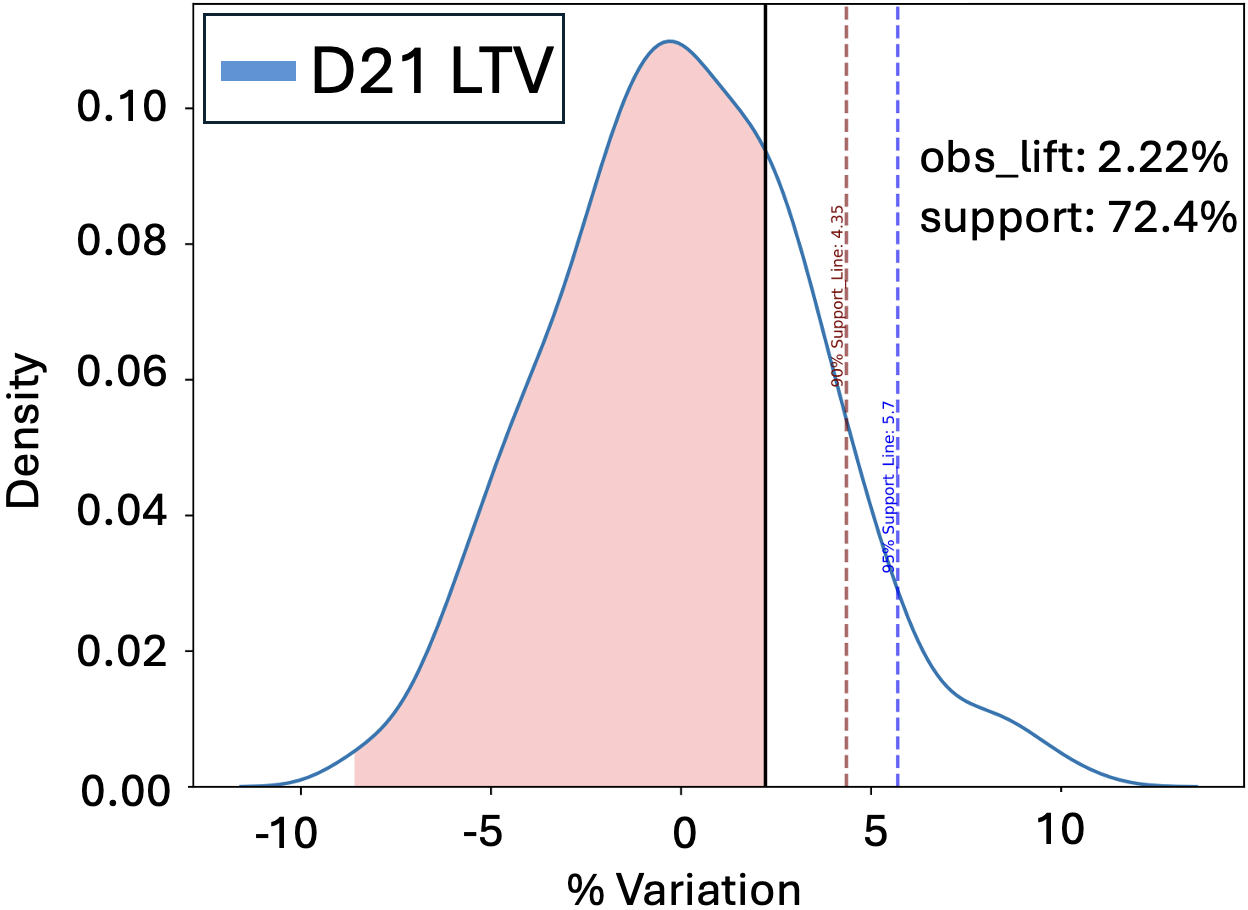}     
    }
    \subcaptionbox{Diffusion-QL \label{fig_AB_diffusion}}{
        \includegraphics[scale=0.118]{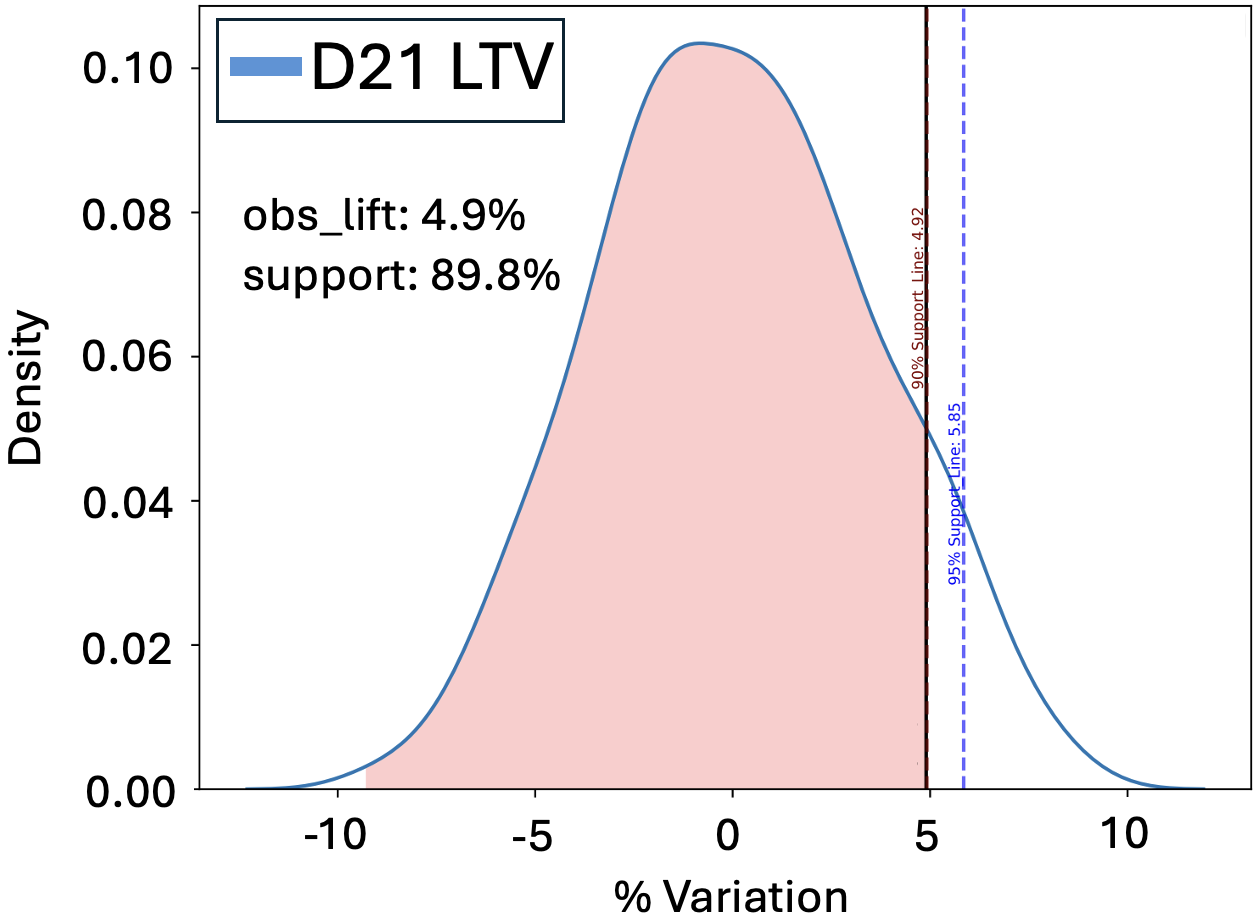}   
    }
    \caption{\textmd{D21 LTV lift of FAST-Q w.r.t Diffusion, TD3+BC and old policies using our A/B test conclusion framework. The red and blue dotted vertical lines signify 90\% and 95\% support respectively. }}
    \label{fig_AB_plots}
\end{figure}

Figure \ref{fig_AB_plots} shows D21 LTV lifts of players with FAST-Q against the 3 competitive paths with our in-house robust A/B experimentation conclusion framework. It mitigates the challenges associated with skewed and constantly evolving user base. As many treatments are significant only on the top 2-3\% of the users, the conclusions are often susceptible to bias from the statistical outliers, like the deceptive engagement and return time lifts shown by sub-optimal policies. The conclusion framework combines stratified sampling, outlier detection and uses permutation testing to facilitate direct comparisons of the LTV metric without depending on normality assumptions or being sensitive to rank changes, as seen in t-test and the Mann-Whitney test~\cite{mannwhitney}.  A crucial metric is the \textbf{support \%}, which represents the percentage of instances in the null distribution that fall below the observed lift (test statistics). A higher support indicates a greater likelihood/confidence of a positive impact. We see that FAST-Q outperforms all models with net LTV lift (adjusting on the cost against paid rewards) of at least 2\% or higher with  high \textit{support} over the entire 21 days period.


\subsection{Evaluation on open RL datasets}
FAST-Q is also evaluated on D4RL~\cite{fu2021d4rl} datasets of OpenAI gym~\cite{openai_gym} MuJoCo~\cite{mujoco} tasks. We compare FAST-Q with our baseline TD3+BC \cite{minimalist} model. Since each of the dataset corresponds to a single policy, we make slight modifications to our architecture to enable training on such datasets. We recommend the following changes to adapt to single-policy setting: 1) removing PE layer, 2) modifying BR layer training as recommended in the original paper~\cite{ganin}.
Table ~\ref{tab:d4rl_evaluations} shows the average normalized score over the final 10 evaluations and 5 seeds. The $\pm$ captures the standard deviation over seeds. TD3+BC results are same as the ones reported by the authors in~\cite{minimalist}. We observe that FAST-Q outperforms on 6 out of 9 policy-tasks quite significantly whereas misses on two out of the remaining three by narrow margins, indicating a good applicability of the BR training for these tasks as well. 
\begin{table}[h]
    \centering
    \begin{tabular}{lcc}
    \toprule
        \textbf{Gym Task} & \textbf{TD3+BC} & \textbf{Fast-Q} \\ 
    \midrule
        halfcheetah-medium-v0 & \textbf{42.8} & 39.9 $\pm$ 0.25 \\
        hopper-medium-v0 & 99.5 & \textbf{102.2} $\pm$ 2.05 \\
        walker2d-medium-v0 & 79.7 & \textbf{82.4} $\pm$ 2.19 \\
    \midrule
        halfcheetah-medium-replay-v0 & \textbf{43.3} & 42.3 $\pm$ 0.52 \\
        hopper-medium-replay-v0 & 31.4 & \textbf{49.1} $\pm$ 0.40 \\
        walker2d-medium-replay-v0 & 25.2 & \textbf{46.5} $\pm$ 4.47 \\
    \midrule
        halfcheetah-medium-expert-v0 & \textbf{97.9} & 50.9 $\pm$ 2.82 \\
        hopper-medium-expert-v0 & 112.2 & \textbf{114.3} $\pm$ 1.32 \\
        walker2d-medium-expert-v0 & 101.1 & \textbf{110.2} $\pm$ 1.79 \\
    \bottomrule
    \end{tabular}
    \caption{\small \textmd{Performance comparison of FAST-Q with TD3+BC}}
    \label{tab:d4rl_evaluations}
    \vspace{-20pt}
\end{table}

\section{Conclusion and Future Work}
We present FAST-Q which addresses the problem related to  highly sparse, partially overlapping state spaces across policies influenced by the experiment path selection logic which biases state spaces towards specific policies. Current SOTA methods like TD3+BC and Diffuion-QL, constrain learning from such offline data by clipping known
counterfactual actions as out-of-distribution due to poor generalization across unobserved states, a long pending problem, also identified in \cite{minimalist,fujimoto2019} but not been addressed yet. We learned that Open datasets like Gym-MuJoCo tasks do not exhibit these challenges and hence  this problem does not surface as crucial to the research community. FAST-Q while proposing the \textbf{Balanced Representation} (BR), \textbf{Q-value decomposition} for multi-objective learning and \textbf{Fast-track exploration} on the offline data demonstrates \textit{at least} \textbf{0.15\%} increase in player returns, \textbf{2\%} improvement in lifetime value (LTV), \textbf{0.4\%} enhancement in the recommendation driven engagement, \textbf{2\%} improvement in the players' platform dwell time and an impressive \textbf{10\%} reduction in the costs associated with the recommendation, on our volatile gaming platform.

Further to this, FAST-Q also shows highly promising results on Gym-MuJoCo tasks with state generalization via BR training. Though FAST-Q leverages TD3+BC for policy regularization, we believe that moving ahead, we can also leverage more recent and better performing technique suggested in Diffusion-QL for policy regularization. However, choice of TD3+BC was primarily due to it being established as a minimalist but a highly effective approach for policy regularization~\cite{minimalist}.

\bibliographystyle{plain}


\newpage
\begin{appendices}
\section{Algorithm}
\label{sec:appendix}
\subsection{FAST-Q Algorithm}
\label{app:algorithm}
\begin{algorithm}
\caption{Q-learning with Policy Expert and Balancing Representation}
\begin{algorithmic}[1]
\State Initialize policy network $\pi_\theta$, critic network $Q_{\phi}$, target networks $\pi_{\theta'}$, $Q_{\phi'}$, Policy Expert $\beta_{p_{n}}$, Balancing Representation function $\Theta$ and exploration factor $\epsilon$
\For{each epoch}
    \State Sample mini-batch $D = \{(S_t, A_t, R_t, S_{t+1})\} \sim \mathcal{B}$
    \State \textbf{\textit{Critic Update with Policy Expert (PE) and Balancing Representation (BR)}}
    \State $S_t^{PE} = \beta_{p_n}(S_t)$
    \State $S_t^{BR} = \Theta(S_t^{PE})$
    \State Compute Critic target:
    \[
        y_t = R_t + \gamma \max_{a'} Q_{\phi'}(S_{t+1}^{BR}, \pi_{\theta'}(S_{t+1}^{BR}))
    \]
    \State Compute Critic Q-value loss:
    \[
        \mathcal{L}_{\text{critic}} = \frac{1}{|D|} \sum_{t} \left( Q_{\phi}(S_t^{BR}, A_t) - y_t \right)^2
    \]
    \State Compute MSE between weighted Q-value and reward:
    \[
        \mathcal{L}_{\text{MSE}} = \frac{1}{|D|} \sum_{t} \left( w_t Q_{\phi}(S_t^{BR}, A_t) - R_t \right)^2
    \]
    \State \textbf{\textit{Actor Update: Adjust action selection with exploration}}
    \State Sample exploration probability $e \sim \text{Uniform}(0, 1)$
    \If{$e \geq \epsilon$}
        \State Select action $a = A_t$ (Greedy action based on current policy)
    \Else
        \State Select action $a = \hat{A}_\text{cp}$ (Counterfactual policy (CP) action)
    \EndIf
    \State Update policy:
    \[
        \pi = \arg \max_{\pi} \, \mathbb{E} \left[ \lambda \, Q(S_t^{BR}, \pi(S_t^{BR})) - (\pi(S_t^{BR}) - a)^2 \right]
    \]
    \State \textbf{\textit{Update target networks}}
    \State $\theta' \gets \tau \theta + (1 - \tau) \theta'$
    \State $\phi' \gets \tau \phi + (1 - \tau) \phi'$
\EndFor
\end{algorithmic}
\end{algorithm}

\subsection{Hyperparameters}
\label{app:hyperparameters}
Table~\ref{tab:hyperparameters} lists down the hyperparameters used by FAST-Q.
\begin{table}[h]
    \centering
    \small
    \begin{tabular}{lc}
    \toprule
        \textbf{Hyperparameter} & \textbf{Value} \\ 
    \midrule
        Discount factor ($\gamma$) & 0.7 \\
        Target update rate ($\tau$) & 0.005 \\
        $\alpha$~\cite{minimalist} & 2.5 \\
        Optimizer & Adam~\cite{adam_optimizer} \\
        Learning rate & 3e-4 \\
        Batch size & 1024 \\
        Policy noise & 0.2 \\
        Policy noise clipping & (-0.5, 0.5) \\
        Policy update frequency & 2 \\
    \bottomrule
    \end{tabular}
    \caption{\textmd{FAST-Q Hyperparameters}}
    \label{tab:hyperparameters}
\end{table}

\subsection{User volatility on RummyCircle platform}
\label{app:user_volatility}

\begin{figure}
  \includegraphics[width=0.45\textwidth, height=4cm]{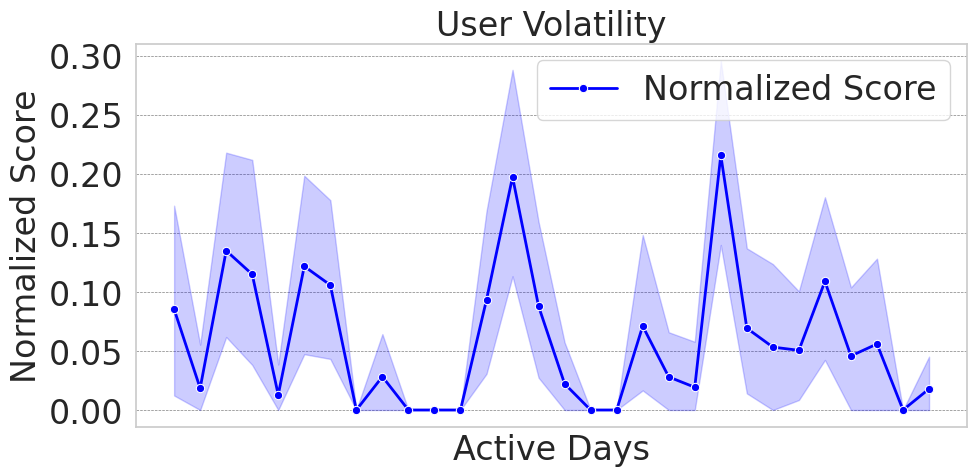}
  \caption{\textmd{Volatility in Player Enagement on our platform}}
  \label{fig_user_volatility}
\end{figure}

Our platform experiences many challenges such as skewed user base, the influence of top percentile players, and constantly evolving user behavior resulting in shifting user patterns and trends. Many treatments primarily affect the top 2-3\% of users, making measurements susceptible to noise. This complicates distinguishing between statistical outliers and genuine high-value players, heightening the risk of false positives or prematurely dismissing promising ideas. Figure \ref{fig_user_volatility} highlights as a representative example volatility in Player engagement on our platform over time. We see  uncertainties with shifts in the mean values over the entire time period.  

\subsection{Details on Hopper and Player Data State Representation:}
\label{appendix:sparsity}
In the Figure \ref{fig_state_isolation_across_policy_classes} The state representation for Hopper is directly obtained by 2-D mapping using UMAP ~\cite{umap}, whereas the time-variant player state information was mapped using an auto-encoder. Secondly for the Figure \ref{fig_state_isolation_sparsity}, a density-based clustering algorithm, OPTICS ~\cite{OPTICS}, was applied to both the datasets, with variations in the minimum number of samples in a neighborhood required for a point to be considered a core point. Our data consistently exhibited a greater number of distinct clusters, indicating more sparse data distribution.


\end{appendices}
\end{document}